\documentclass[acmsmall]{acmart}
\usepackage[utf8]{inputenc} % allow utf-8 input
\usepackage[T1]{fontenc}    % use 8-bit T1 fonts
\usepackage{hyperref}       % hyperlinks
\usepackage{url}            % simple URL typesetting
\usepackage{booktabs}       % professional-quality tables
\usepackage{amsfonts}       % blackboard math symbols
\usepackage{nicefrac}       % compact symbols for 1/2, etc.
\usepackage{microtype}      % microtypography
\usepackage{lipsum}

\usepackage{amsmath,amsfonts}

\usepackage{graphicx}
\usepackage{textcomp}
\usepackage{xcolor}
\usepackage{fancyhdr}

\usepackage{graphicx}
\usepackage{amsmath}
\usepackage{epstopdf}
\usepackage{multirow}
\usepackage{algorithm}
\usepackage{algorithmicx}
\usepackage{algpseudocode}
\usepackage{hhline}
\usepackage{enumitem}
\usepackage{listings}
\usepackage{anyfontsize}
\usepackage{wrapfig}
\usepackage{footnote}
\usepackage{tablefootnote}
\usepackage{wrapfig}
\usepackage{soul}
\usepackage{tikz}

\makesavenoteenv{tabular}
\makesavenoteenv{table}
\graphicspath{ {figures/} }
\usepackage{array}
\newcolumntype{L}[1]{>{\raggedright\arraybackslash}p{#1}}
\newcolumntype{C}[1]{>{\centering\arraybackslash}p{#1}}
\newcolumntype{R}[1]{>{\raggedleft\arraybackslash}p{#1}}

\newcolumntype{?}{!{\vrule width 1pt}}

\usepackage{xcolor,colortbl}

\newcolumntype{y}{>{\columncolor{yellow}}c}

\usepackage{fancyhdr}
\usepackage[normalem]{ulem}

\usepackage{boldline}

\newcommand{\thename}{GANDSE}

\setcopyright{acmcopyright}
\acmJournal{TODAES}
\acmYear{2022} \acmVolume{1} \acmNumber{1} \acmArticle{1} \acmMonth{1} \acmPrice{15.00}\acmDOI{10.1145/3570926}

\begin{document}

%%
%% The "title" command has an optional parameter,
%% allowing the author to define a "short title" to be used in page headers.
\title{\thename{}: \underline{G}enerative \underline{A}dversarial \underline{N}etwork
based \underline{D}esign \underline{S}pace \underline{E}xploration for Neural Network Accelerator Design}

%%
%% The "author" command and its associated commands are used to define
%% the authors and their affiliations.
%% Of note is the shared affiliation of the first two authors, and the
%% "authornote" and "authornotemark" commands
%% used to denote shared contribution to the research.

\author{Lang Feng}
%\authornotemark[1]
\email{flang@nju.edu.cn}

\affiliation{%
  \institution{Nanjing University}
  \streetaddress{163 Xianlin Road}
  %\city{Qixia District}
  \state{Nanjing}
  \country{China}
  \postcode{210023}
}

\author{Wenjian Liu}
%\authornotemark[1]
\email{wjliu@nju.edu.cn}

\affiliation{%
  \institution{Nanjing University}
  %\streetaddress{163 Xianlin Road}
  %\city{Qixia District}
  %\state{Nanjing}
  \country{China}
  %\postcode{210023}
}

\author{Chuliang Guo}
%\authornotemark[1]
\email{chuliang007@zju.edu.cn}

\affiliation{%
  \institution{Zhejiang University}
  \streetaddress{866 Yuhangtang Road}
  \state{Hangzhou}
  \country{China}
  \postcode{310058}
}

\author{Ke Tang}
%\authornotemark[1]
\email{mf21230105@smail.nju.edu.cn}

\affiliation{%
  \institution{Nanjing University}
  %\streetaddress{163 Xianlin Road}
  %\city{Qixia District}
  %\state{Nanjing}
  \country{China}
  %\postcode{210023}
}

\author{Cheng Zhuo}
\authornotemark[1]
\email{czhuo@zju.edu.cn}

\affiliation{%
  \institution{Zhejiang University}
  %\streetaddress{866 Yuhangtang Road}
  %\state{Hangzhou}
  \country{China}
  %\postcode{310058}
}

\affiliation{%
  \institution{Key Laboratory of Collaborative Sensing and Autonomous Unmanned Systems of Zhejiang Province}
  %\streetaddress{866 Yuhangtang Road}
  %\state{Hangzhou}
  \country{China}
  %\postcode{310058}
}

\author{Zhongfeng Wang}
\authornotemark[1]
\email{zfwang@nju.edu.cn}

\affiliation{%
  \institution{Nanjing University}
  %\streetaddress{163 Xianlin Road}
  %\city{Qixia District}
  %\state{Nanjing}
  \country{China}
  %\postcode{210023}
}

%%
%% By default, the full list of authors will be used in the page
%% headers. Often, this list is too long, and will overlap
%% other information printed in the page headers. This command allows
%% the author to define a more concise list
%% of authors' names for this purpose.
%\renewcommand{\shortauthors}{Trovato and Tobin, et al.}

%%
%% The abstract is a short summary of the work to be presented in the
%% article.
\begin{abstract}
With the popularity of deep learning, the hardware implementation platform of deep learning has received increasing interest. Unlike the general purpose devices, $e.g.$, CPU, or GPU, where the deep learning algorithms are executed at the software level, neural network hardware accelerators directly execute the algorithms to achieve higher both energy efficiency and performance improvements.
However, as the deep learning algorithms evolve frequently, the engineering effort and cost of designing the hardware accelerators are greatly increased. To improve the design quality while saving the cost, design automation for neural network accelerators was proposed, where design space exploration algorithms are used to automatically search the optimized accelerator design within a design space. Nevertheless, the increasing complexity of the neural network accelerators brings the increasing dimensions to the design space. As a result, the previous design space exploration algorithms are no longer effective enough to find an optimized design. In this work, we propose a neural network accelerator design automation framework named \thename{}, where we rethink the problem of design space exploration, and propose a novel approach based on the generative adversarial network (GAN) to support an optimized exploration for high dimension large design space. The experiments show that \thename{} is able to find the more optimized designs in negligible time compared with approaches including multilayer perceptron and deep reinforcement learning.
\end{abstract}

%%
%% The code below is generated by the tool at http://dl.acm.org/ccs.cfm.
%% Please copy and paste the code instead of the example below.
%%


%%
%% Keywords. The author(s) should pick words that accurately describe
%% the work being presented. Separate the keywords with commas.

%%
%% This command processes the author and affiliation and title
%% information and builds the first part of the formatted document.
\maketitle
\newcommand\blfootnote[1]{%
\begingroup
\renewcommand\thefootnote{}\footnote{#1}%
\addtocounter{footnote}{-1}%
\endgroup
}
\blfootnote{$^*$The corresponding authors.}

\section{Introduction}

Deep learning is prevalent in the area of artificial intelligence (AI) due to its success in many fields such as vision and speech processing. 
Initially, the training and inference of deep learning were deployed on the central processing unit (CPU). After exploring the parallel computing capability of the graphics processing unit (GPU), GPU became the most common platform for deep learning \cite{ChetlurWVCTCS14}. 
However, with the growth of the deep learning model size, using GPU can incur large power consumption that is not acceptable for many low power applications, such as edge computing.

To control the power consumption while maintaining the performance of executing large neural network models, the dedicated hardware neural network accelerators are proposed, such as Eyeriss~\cite{Chen17}, Gemmini~\cite{Genc19}, DaDianNao{~\cite{Luo17}}, etc. 
However, deep learning algorithms keep updating frequently. As designing the hardware accelerators requires a long period, frequent algorithm updating can introduce significant engineering effort and cost. To avoid this, researchers try to seek the design automation approaches for neural network accelerators. 

There are various studies of neural network accelerator design automation, such as DnnWeaver \cite{Sharma16}, DNNBuilder~\cite{Zhang18}, AutoDNNchip~\cite{Xu20}, ConfuciuX~\cite{Kao20}, etc. Each of them proposed an automatic design framework to search and generate the optimized accelerator, given a deep learning inference task. 
Given the requirements from the user, the searching problem to get the optimized design that satisfies the requirements is called the design space exploration (DSE) problem, which is a step in neural network accelerator design automation. 
However, each of the previous DSE algorithm~\cite{Sharma16,Zhang18,Abdelouahab16,Xu20,Zhang20,Lin21,Sohrabizadeh21,Kao20} iteratively uses a model to evaluate the quality of the searching result, and use a searching algorithm to update the searching result given the evaluation. The iterative searching can incur large searching time cost. Recent study~\cite{Samajdar21} proposes a non-iterative approach by using multilayer perceptron (MLP) for DSE, but our experiments show that MLP is still not competent under high dimension design space. Therefore, a new algorithm that is able to effectively search the optimized design in a high dimension large design space is needed.

The neural network accelerator design automation is to generate a neural network accelerator, given the objectives from the designers. This is similar to a kind of neural networks called generative adversarial networks (GANs), which are usually used to generate images that satisfy the requirements from the users. Inspired by this, we investigate the potential to apply GAN to DSE tasks, and propose a GAN-based DSE flow.
Because GAN usually has millions of outputs for image pixels, when using GAN for outputting the configurations, it has the potential to support high dimension large design space. Besides, the iterative searching is also avoided.

In our work, we propose \thename{}, which is a neural network accelerator design automation framework with the GAN-based DSE. Different from previous work, \thename{} introduces a new DSE flow that can effectively find more optimized design within a high dimension large design space. To the best of our knowledge, this is the first work that explores the potential to applying GAN in the field of DSE.
The detailed contributions of this work are in the following.
\begin{itemize}[leftmargin=2ex]
    \item A neural network accelerator design automation framework named \thename{} is proposed to automatically generate the synthesizable RTL code of the optimized neural network accelerator design given a deep learning task from the user. The generated code can be implemented on an FPGA. 
    \item A new DSE flow based on GAN is proposed, which provides a new direction for handling DSE problems, and the experiment results indicate the effectiveness. 
    \item To fit our neural network accelerator design automation problem, the modifications are proposed to the structure of GAN, the training flow, the loss function design, the encoding of the features, and the candidate selection after the inference, etc. 
    \item Throughout experiments are performed, which prove the effectiveness of the DSE algorithm in \thename{}. Compared with other approaches including simulated annealing, multilayer perceptron, and deep reinforcement learning, \thename{} finds more designs that satisfy the user's requirements in negligible time. %Besides, we evaluate the synthesized designs of different DSE algorithms.
\end{itemize}

In the following sections, Section~\ref{sec:background} introduces the background of design space exploration and generative adversarial networks. Section~\ref{sec:prev_work} discusses the related work. Section~\ref{sec:motivation} introduces the motivation and design rationale of \thename{}. The overview of \thename{} is introduced in Section~\ref{sec:system}, and the design details are then proposed in Section~\ref{sec:detail}. After that, the experimental results are analyzed in Section~\ref{sec:exp}. Finally, Section~\ref{sec:conclusion} concludes the paper.

\section{Background}
\label{sec:background}

\subsection{Ordinary Neural Network Accelerator Design Space Exploration}
\label{sec:dse}

Design space exploration (DSE) is the approach to search for the design, which satisfies the given \textbf{objectives}, in the design space. For example, the DSE discussed in this work is searching for a set of \textbf{configurations} of the neural network accelerators (such as the number of the processing elements (PEs), the size of SRAM, etc.), which makes the accelerators satisfy some objectives such as the requirements to the latency, the power consumption, etc. 
%For the examples of the current section, the design space is composed of all the possible configurations of PE numbers and SRAM sizes. 

\begin{figure}[!hbt]
	\centering
	%\vspace{0.1in}
	%\vspace{-1ex}
	%\includegraphics[width=\columnwidth]{figures/cyclebd.eps}
	\includegraphics[width=0.75\columnwidth]{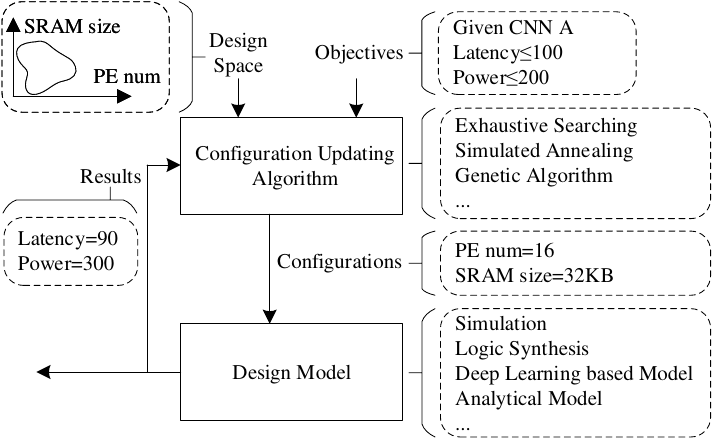}
	\caption{The ordinary DSE flow and examples of neural network accelerator design automation.}
	\label{fig:dseflow}
	%\vspace{-3ex}
\end{figure}

According to the survey~\cite{Schafer20}, most of the DSE algorithms can be illustrated in Figure~\ref{fig:dseflow}. Two critical components are the \textbf{configuration updating algorithm} and the \textbf{design model}. 
The design model is a model of the metrics in the objectives.
Given a set of configurations, the design model is able to output the values of the metrics (Note that for easy of discussion, in the following paper, the values of the metrics in the objectives such as ``Latency = 90 cycles'' are also referred as ``objectives'').
The design model can be constructed in various ways, including the analytical model, the deep neural networks, etc.
For the configuration updating algorithm, it is to output a set of the configurations, given the design space, the user's objectives, and the latest results of the metrics in the objectives. Some typical algorithms are exhaustive searching, simulated annealing, genetic algorithms, etc.

However, the DSE flow in Figure~\ref{fig:dseflow} greatly depends on iteratively searching the optimized design, which is relatively less effective. In contrast, to avoid the iterative searching and explore the new direction for DSE, we propose a completely new DSE flow based on GAN in \thename{}.

\subsection{Generative Adversarial Networks}

Generative Adversarial Networks (GANs)~\cite{goodfellow14} are the neural networks that are able to generate new data, which have the same distribution as the training data. Specifically, conditional GANs (cGANs)~\cite{Mirza14} can generate the new data given some specifications, which are called the conditional information. For example, a cGAN can be trained to  generate new images of specific kinds of animals with specific colors required by the users. 
Two critical neural networks involved in cGAN are called the \textbf{generator (G)} and the \textbf{discriminator (D)}. The generator can generate the new data while the discriminator can tell if the data is real (from the training set) or fake (generated from the generator). 

\begin{figure}[!hbt]
	\centering
	%\vspace{0.1in}
	%\vspace{-1ex}
	%\includegraphics[width=\columnwidth]{figures/cyclebd.eps}
	\includegraphics[width=0.75\columnwidth]{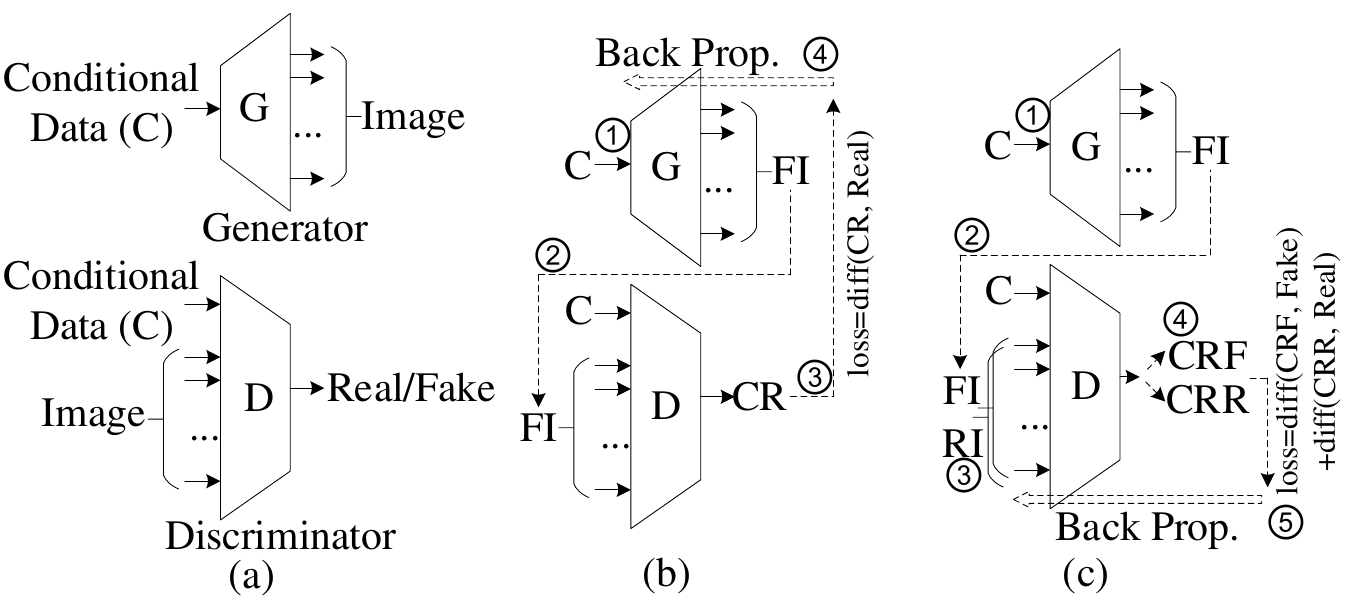}
	\caption{(a) An example of a cGAN's architecture; (b) The training procedures of G; (c) The training procedures of D. (Note that G has an additional input with small random numbers as noise, which is not shown in this figure.)}
	\label{fig:ganex}
	\vspace{-1ex}
\end{figure}

Figure~\ref{fig:ganex} shows an example of using cGAN for image generation. Given the conditional data, G tries to generate an image satisfying the condition and confuse D, while the goal of D is to discriminate if an image is a real image or a fake image. 
When training G, G is given the conditional data (C) of the images in the training set, and outputs the generated fake images (FI) accordingly. The fake images are then input to D, and the output classification results (CR) indicate if the images are real or not. The loss of G is to measure the difference between CR and ``Real'', since it means G can generate the images that seem real enough if D classifies FI as ``Real''. This loss is used to do the back propagation of G. When training D, 
the real images (RI) from the training set and fake images (FI) generated from G are input to D. Then, D will output the classification results CRR and CRF on the real images and the fake images, respectively.
The loss of D is to measure the difference between CRF and ``Fake'', and that between CRR and ``Real'', which represent the precision of the classification. Then, this loss is used for the back propagation of D.
As the relationship between the image and the conditional data is similar to that between the  accelerator configurations and the user's objectives, cGAN might also be used for the DSE.
Because cGAN usually has millions of outputs for pixels, when using cGAN for outputting the configurations, it has the potential to support high dimension large design space.
Besides, since only one-time inference for each configuration is needed, it may save the time cost compared with running the iterative DSE algorithms.

\section{Related Work}
\label{sec:prev_work}

With the development of neural networks, many neural network accelerators are also proposed. For example, Eyeriss{~\cite{Chen17}} was designed for the energy efficient inference of convolutional neural networks (CNNs), with its proposed dataflow.
NVDLA{~\cite{NVDLA}} is a more general neural network accelerator that is configurable and supports IoT devices. 
Gemmini{~\cite{Genc19}} can generate the systolic arrays for matrix multiplication, and it can be integrated into RISC-V processors.
DaDianNao{~\cite{Luo17}} is a multi-chip machine-learning architecture for realizing high-degree parallelism of neural network inference and training.
Besides, Cambricon-Q{~\cite{Zhao21}} was also proposed for neural network training. It supports the efficient quantized training, and also has negligible accuracy loss.
These proposed accelerators are manually well-designed, but the algorithms for neural
networks evolve quickly, and manually designing the hardware accelerators can introduce significant engineering effort and cost. Different from these studies, the purpose of \thename{} is to automatically generate the accelerators that perform well enough to reduce the design cost.

There have been various studies about neural network accelerator design automation. Two typical design frameworks are DnnWeaver \cite{Sharma16} and DNNBuilder~\cite{Zhang18}. DnnWeaver proposes an end-to-end design framework for generating the RTL code of the neural network accelerator, given the deep neural network (DNN) model written by the users using a proposed domain-specific language. The neural network accelerator is generated by setting the parameters of a Verilog template. DNNBuilder is similar to DnnWeaver, except that the template of DNNBuilder contains multiple systolic arrays, each of which processes one layer of the calculation of DNN inference.
Meanwhile, AutoDNNchip~\cite{Xu20} provides multiple detailed IP templates and global architectures that are composed of the chosen IPs. It expands the design space with higher granularity.
DNNExplorer~\cite{Zhang20} uses a divide and conquer method for the DSE.
In work~\cite{Zeng2018}, a fast DSE algorithm is proposed for searching the optimized size of the FFT for CNN calculations. The optimized size is found by calculating the intersections of the curves of the communication bound and computation bound. 
To avoid the iterative searching, AIRCHITECT~\cite{Samajdar21} uses multilayer perceptron (MLP) for generating the accelerator design. However, our experiments show that only applying MLP finds limited optimized results given high dimension large design space.
ConfuciuX~\cite{Kao20} uses reinforcement learning and the genetic algorithm for searching the optimized dataflow and hardware resource assignment.  
%Compared with the studies above, recent study 
NAAS~\cite{Lin21} proposes a design framework integrating the searches for the neural network architectures, the mapping methods, and the neural network accelerators. 

Besides the approaches using the templates, using high level synthesis (HLS) tools can also generate the neural network accelerators. AutoDSE~\cite{Sohrabizadeh20} proposes a DSE approach that uses the bottleneck-guided gradient optimizer to find the bottleneck of the design and optimize it accordingly. The work~\cite{Abdelouahab16} optimizes the HLS configurations with the use of exhaustive searching as the DSE algorithm.
%Some other works focus on improving the design model in DSE. 
The work~\cite{Sohrabizadeh21} uses the graph neural network (GNN) to increase the precision of the design model, and thus helps the quality of the DSE. 

However, the DSE algorithms used in the previous studies related to neural network accelerator design automation either follow the ordinary DSE flow in Figure~\ref{fig:dseflow} or have low dimension design space, so they cannot effectively find the accelerator design that is optimized enough with the growing of  the design space dimensions.

\section{Motivation and Design Rationale}
\label{sec:motivation}

In this section, we introduce the motivation and the core design rationale of \thename{}. 
The goal of the DSE for neural network accelerator design automation is that, given the user's objectives and the neural networks to be executed, a set of the configurations of the neural network accelerator needs to be generated, so that the configured neural network accelerator can satisfy the objectives. 

\begin{figure*}[!t]
    \centering
    \includegraphics[width=1\columnwidth]{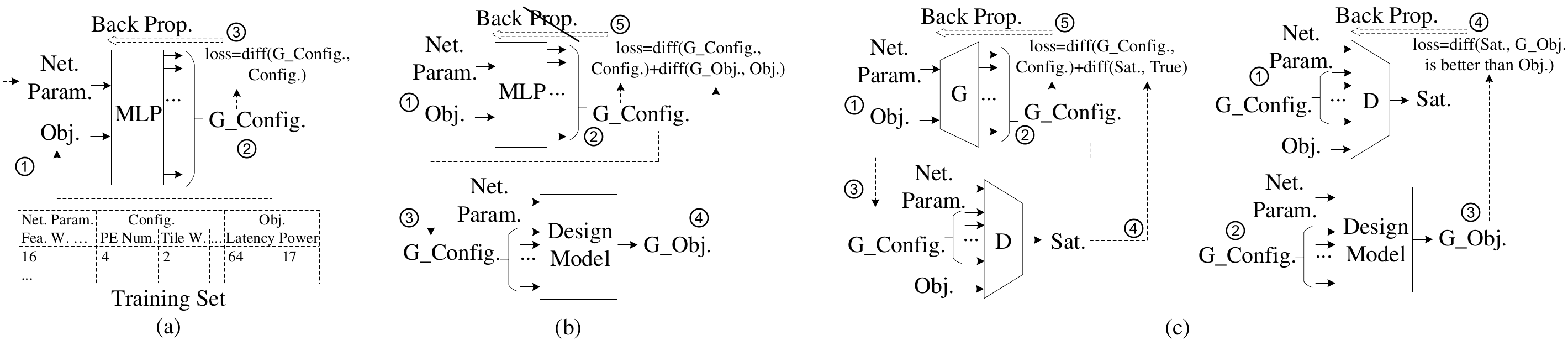}
    \caption{The training flows in different DSE schemes: (a) Barely using MLP; (b) Using MLP with the design model for improving the loss function, which is not viable; (c) Using GAN, where the left and right parts are for training the G and the D, respectively. (``Net. Param.'', ``Obj.'', ``G\_Config.'', ``G\_Obj.'' stand for network parameters, objectives, generated configurations, objectives under the generated configurations, respectively.)}
    %\vspace{-2ex}
\label{fig:train}
\end{figure*}

Basically, to avoid the iterative searching of the ordinary DSE algorithms in Figure~\ref{fig:dseflow},  similar as work~\cite{Samajdar21}, the DSE can be realized by the basic neural networks named multilayer perceptron (MLP), where the inputs are the objectives and the \textbf{network parameters} that interpret the neural network architectures, and the outputs are the configurations.
The training flow of MLP is shown in Figure~\ref{fig:train}(a). 
First, the training set is generated by evenly choosing enough sets of the network parameters and configurations, and using the design model for the corresponding objectives.
Then, by using the difference between the configurations in the training set and the generated configurations as the loss, the MLP is trained to learn the distribution of the data in the training set. 
Ideally, the MLP is able to generate the satisfied configurations during the inference.

However, there is a critical challenge in the approach in Figure~\ref{fig:train}(a). That is, if the generated configurations differ from the corresponding configurations in the training set, it is regarded as a wrong set of configurations since it incurs the loss. Nevertheless, there are usually multiple sets of the configurations that can satisfy the objectives for the given network parameters. Although the generated configurations are different from those in the training set, it is still possible that the generated configurations are also satisfied, or even better satisfied the objectives. The naive approach in Figure~\ref{fig:train}(a) ignores this scenario, so it can miss many opportunities to generate better configurations.

To try to address this problem, a direct approach is to use the design model to verify if the generated configurations satisfy the objectives, when the generated configurations are different from the training set. The training flow is in Figure~\ref{fig:train}(b). However, the loss function contains the corresponding objectives ``G\_Obj'' by the generated configurations, which relates to the weights of the MLP and the design model. This means the gradients of MLP also relate to the design model. As the design model is not necessarily analytic, it means the loss is not necessarily able to be back propagated. Therefore, the approach in Figure~\ref{fig:train}(b) is not viable.

Finally, inspired by using the GAN (specifically, cGAN) for the text-image problem, where the scenario is similar (creating the data that satisfy the objectives), we proposed a GAN-based DSE algorithm. The training flow of the proposed GAN is shown in Figure~\ref{fig:train}(c).
Compared with Figure~\ref{fig:train}(b), now the MLP is called the generator (G), and the design model is replaced by the discriminator (D). The D outputs if the configurations can satisfy the objectives given the network parameters, which is called ``satisfaction'' (``Sat.'' in Figure~\ref{fig:train}(c)). In this case, since the calculations in both G and D are differentiable, the loss can be back propagated to train G. When training D, the generated configurations from G are input to D to get the satisfaction prediction. Besides, the corresponding objectives ``G\_Obj'' of the same generated configurations are obtained from the design model. If ``G\_Obj'' is overall better than the user's objectives, the actual satisfaction should be ``True'', otherwise, it is ``False''. Then the actual satisfaction is compared with the satisfaction prediction to get the loss. Note that although the actual satisfaction in the loss function relates to the design model, it does not relate to the weight in D. This means, the actual satisfaction is a constant for calculating the gradients of D, and thus, the same problem in Figure~\ref{fig:train}(b) does not exist in this case. The G and D are trained alternatively based on each other's outputs, and thus the adversarial training is realized.

\section{System Overview}
\label{sec:system}

\begin{figure*}[!t]
    \centering
    \includegraphics[width=0.97\columnwidth]{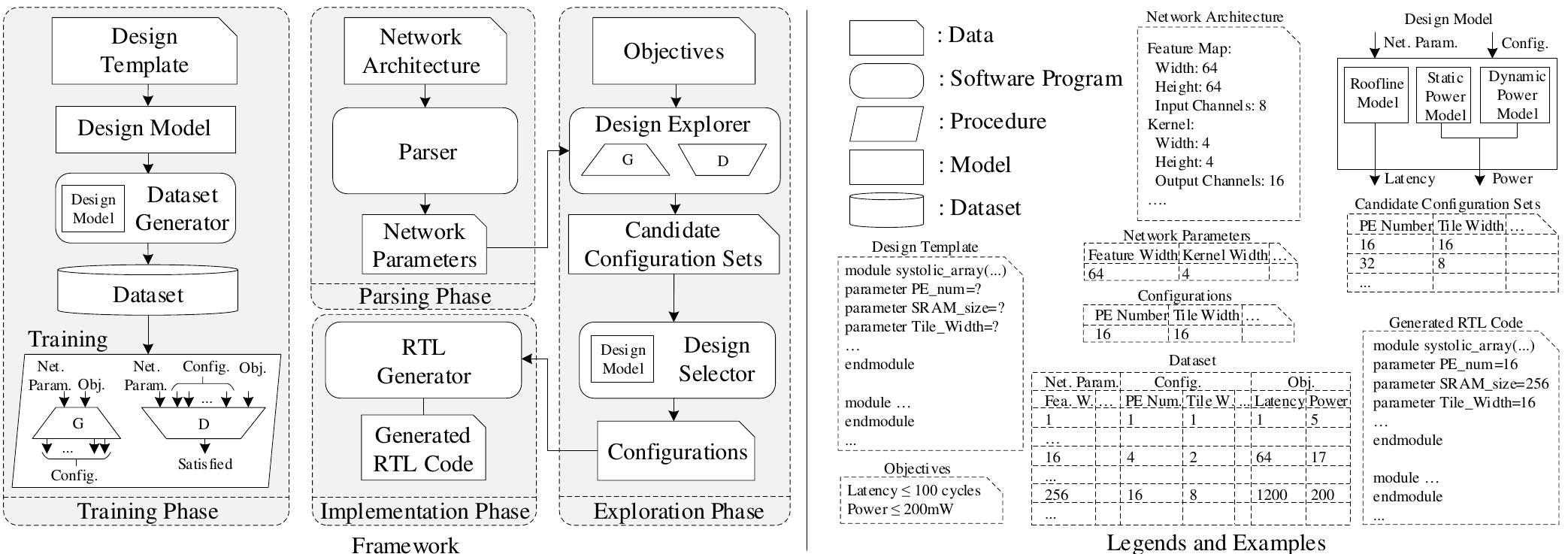}
    \caption{The overview of \thename{} framework. (The training phase only needs to be performed once for each design template. Other 3 phases need to be performed for each set of network architectures and objectives.)}
    \vspace{-3ex}
\label{SystemOverview}
\end{figure*}

Based on the discussion in Section~\ref{sec:motivation}, an end-to-end neural network accelerator design automation framework called \thename{} is proposed to search for the highly optimized design in a high dimension large design space with small time cost, where a novel DSE flow is designed. 
\thename{} can generate the optimized neural network accelerator, which can be used for the inference of the given neural networks while satisfying the user's objectives of the latency and the power consumption. 
The framework of \thename{} is shown in Figure~\ref{SystemOverview}, with three inputs and four phases involved. 
%The framework can be separated into online and offline stages. 

The inputs are listed in the following.
\vspace{-1ex}
\begin{itemize}[leftmargin=1em]
    \item \textbf{Design Template:} The neural network accelerator design template with a set of configurations needed to be determined, which is written in synthesizable RTL code. The configurations not only contain the \textbf{architecture parameters} such as the SRAM size, but also contain the \textbf{mapping strategies} such as the tiling size of the neural networks.
    \item \textbf{Network Architecture:} The layer architecture of the neural networks that need to be executed 
    on the generated neural network accelerator design. An example is shown in Figure~\ref{SystemOverview}.
    \item \textbf{Objectives:} The user's requirements to the generated neural network accelerator when performing the inference of the given network architecture. Our work uses latency and power as the examples, but the objectives are not limited to them. Besides, the format of the objectives in \thename{} is ``objective$\leq x$'', where $x$ is a number given from the user.
\end{itemize}%\setlength\itemsep{5ex}
%\vspace{-1ex}

The four phases are briefly described in the following. 
\vspace{-1ex}
\begin{itemize}[leftmargin=1em]%[start=1,label=Phase \arabic*,leftmargin=3.8em]%\setlength\itemsep{5ex}
    \item \textbf{Training Phase:} The design explorer in \thename{} is trained to be able to generate the optimized configurations of the template. For each design template, the training phase only needs to be performed for one time.
    \item \textbf{Parsing Phase:} This phase parses user readable inputs to the numbers that can be directly used in the next phase. 
    \item \textbf{Exploration Phase:} This phase generates the optimized configurations, which satisfies the user's objectives.
    \item \textbf{Implementation Phase:} This phase applies the generated configurations to the design template and generates the synthesizable RTL code of the neural network accelerator.
\end{itemize}
After performing the training phase to set up \thename{} according to the given neural network accelerator design template, \thename{} can automatically generate synthesizable RTL code of the neural network accelerator with the configurations, which satisfies the user objectives when executing the neural networks with the given network architecture. The circuits synthesized from the generated RTL code can be implemented on an FPGA.

The details of the four phases are introduced as follows. 

%The offline stage only contains the training phase. 
\subsection{Training Phase}
\label{sec:tphase}

Before \thename{} is able to be used, the critical component, which is the GAN in the \textbf{Design Explorer}, needs to be trained as the flow in Figure~\ref{fig:train}(c).
Note that different from the GNN that is used for the design model in work~\cite{Sohrabizadeh21}, the GAN in the design explorer is for creating new configurations according to the given the objectives, which is the reverse of the design model, and has different challenges discussed in Section~\ref{sec:motivation}.

Given the design template, the design model is created to predict the latency and power. The latency is constructed by using the roofline model, while the power is constructed by combining the static power model and the dynamic power model of the template. The design model is also verified by simulation and synthesis.

After the design model is constructed, an automatic \textbf{Dataset Generator} is employed to generate a dataset containing various neural network architectures, configurations, and the latency and power consumption. The number of the data in the dataset is large enough so that the data can evenly cover the design space. An example of the generated dataset is shown in Figure~\ref{SystemOverview}.

Next, the GAN inside the design explorer is trained with the dataset, and then applied to the design explorer. Since the generator of the GAN in the design explorer learns the mapping from the network parameters and objectives to the configurations through the adversarial learning, it is able to generate the configurations given the user's objectives in the exploration phase. 

%\subsection{Online Stage}

\subsection{Parsing Phase}

In the parsing phase, the \textbf{Network Parser} can parse the user's description that contains the information of the neural network architectures, then generate the \textbf{network parameters} that can be directly input to the next phase along with the objectives. 

The user's description is an abstract intermediate representation for the neural network architectures, such as the network architecture example in Figure~\ref{SystemOverview}. It can be handwritten or extracted from deep learning frameworks such as Caffe~\cite{caffe} or PyTorch~\cite{pytorch}.

\subsection{Exploration Phase}
With the outputs of the parsing phase, the exploration phase can generate the optimized configurations of the neural network accelerator design template that satisfy the user's latency and power objectives.
%By taking the outputs of the parsing phase, 
The \textbf{Design Explorer} contains a GAN, which is trained in the offline phase. 
Based on the generator networks in the GAN, the design explorer tries to generate multiple sets of the configurations including the architecture parameters and the mapping strategies, so that when the configured neural network accelerators are implemented on the FPGA to execute the neural networks with the given network parameters, the latency and power objectives can be satisfied. These sets of the configurations are called \textbf{Candidate Configuration Sets}.
After the candidate configuration sets are generated, the \textbf{Design Selector} then selects the most optimized set of configurations with the use of the selection algorithm, which will be introduced in Section~\ref{sec:detail}.

\subsection{Implementation Phase}
Using the design template with the optimized configurations including the architecture parameters and the mapping strategies, the \textbf{RTL Generator} can generate the synthesizable RTL code, which can be implemented on FPGA boards directly.

\section{GAN-based Design Space Exploration}
\label{sec:detail}

In this section, the critical part of \thename{}, which is the GAN-based DSE algorithm, is further elaborated. The DSE algorithm relates to the training of the GAN, the design explorer (the inference of the GAN), and the design selector in Figure~\ref{SystemOverview}. 

\subsection{Training and Inference Schemes for GAN-based DSE}

The details of the training flow proposed in Section~\ref{sec:motivation} are further introduced.
%A basic GAN-style architecture includes a generator and a discriminator. 
The proposed training scheme is shown in Algorithm~\ref{alg:gantrain}, during which the generator (G) and the discriminator (D) are trained alternately. 
The expectation is, G can output the configurations that can satisfy the objectives for given the network parameters, when the configured neural network accelerator is implemented on the FPGA.
In contrast, the D takes network parameters, configurations, and user's objectives as the inputs. 
The D will decide if the objectives can be satisfied or not and outputs the satisfaction as ``True'' or ``False'', respectively. Note that if and only if all the objectives are better than the user's requirements, the satisfaction is ``True''.
%In the proposed \thename{} framework in Figure~\ref{SystemOverview}, the G takes the user's objectives and the parsed network parameters as the input to generate the configurations including the architecture parameters and mapping strategies, as shown in Figure~\ref{fig:train}(c).
The configurations are one-hot encoded, since most of the configurations of the architectures and mapping strategies are not successive and only some specific numbers are meaningful. Otherwise, the generated configurations might be decimal or negative, which can not be employed.
%Only limited choice will be provided to ensure that all generated architectures and mapping configurations are valid and able to be employed.
The user's objectives and the network parameters are encoded as the binary numbers, and normalized by the standard deviation.
We also employ the one-hot encoding for the satisfaction output of D, similar to other neural networks classification tasks.

\begin{algorithm}[t]
\caption{The proposed training scheme. ($Net_s$: Network parameters of sample $s$; $Config_s$: Configurations of $s$; $LO_s$: Latency objective of $s$; $PO_s$: Power objective of $s$.)} 
\label{alg:gantrain}
\flushleft  {\bf Require:}  
$G$, generator. $D$, discriminator. $Set_{train}$, training set. $iter$, number of iterations. $M_l$, latency model. $M_p$, power model. $E$, cross entropy loss function. $bs$, batch size.
\begin{algorithmic}[1]
\For{$iter$ of training iterations} 
    \For{Each batch $b$ $\in$ $Set_{train}$}
    \State $Loss_{config}, Loss_{critic}, Loss_{dis} \gets 0$
    \For{Each sample $s$ $\in$ $b$}
        \State $Config_g \gets  G(Net_s, LO_s, PO_s)$
        \State $Sat \gets D(Net_s, Config_g, LO_s, PO_s)$
        \State $L_g \gets M_l(N_s, Config_g)$
        \State $P_g \gets M_p(N_s, Config_g, L_g)$
        \State $Loss_{critic} += E(Sat, True)/bs$
        \If{$L_g \leq LO_s$ and $P_g \leq PO_s$}
            \State $Loss_{config} += 0$
            \State $Loss_{dis} += E(Sat, True)/bs$
        \Else
            \State $Loss_{config} += E(Conig_s, Config_g)/bs$
        \State $Loss_{dis} += E(Sat, False)/bs$
        \EndIf
    \EndFor
    \State update $G$ with $Loss_{config}+w_{critic}\times Loss_{critic}$
    \State update $D$ with $Loss_{dis}$
    \EndFor
\EndFor
\State \Return $G$ and $D$
\end{algorithmic}
\end{algorithm}

The detailed training procedures are illustrated in Algorithm~\ref{alg:gantrain}, where we use the mini-batch training (Lines 1-4). 
For each sample $s$ in the training set, the user's objectives $LO_s$ and $PO_s$ and the network parameters $Net_s$ are input to G to get the generated configurations $Config_g$ (Line 5). The generated configurations are then input to D to get the satisfaction $Sat$ (Line 6). 
Next, $Config_g$ are input to the latency model $M_l$ and power model $M_p$ to get the actual latency $L_g$ and power consumption $P_g$ (Lines 7-8), where $M_p$ may need the latency $L_g$ as input of the dynamic power model. 
Then, three losses need to be calculated. The configurations loss $Loss_{config}$ represents if the generated configurations $Config_g$ are close to the actual configurations $Config_s$ of sample $s$. The critic loss $Loss_{critic}$ represents if D believes the generated configurations $Config_g$ satisfies the user's objectives or not. The discriminator loss $Loss_{dis}$ represents if the classification on the satisfaction made by D is correct or not.
$Loss_{critic}$ is calculated by Line 9 by cross entropy loss. The actual latency $L_g$ and power $P_g$ corresponding to the generated configurations are then compared with the objectives $LO_s$ and $PO_s$ of the sample $s$ (Lines 10-16). If $L_g$ and $P_g$ are both better than $LO_s$ and $PO_s$, respectively, $Config_g$ is regarded as the satisfied configurations, so there is no $Loss_{config}$ (Line 11), and the output $Sat$ of D should be ``True'' (Line 12). Otherwise, $Loss_{config}$ is calculated by Line 14, and $Sat$ needs to be false (Line 15). 
After calculating the losses for one batch of the samples, G and D are backpropagated and updated (Lines 18-19). Note that we employ the weight $w_{critic}$ for $Loss_{critic}$ to control the effect of D.

\begin{algorithm}[t]
\caption{The proposed selection algorithm of the design selector.} 
\hspace*{0.02in} {\bf Require:}  
$Set_{config}$, candidate configuration sets. $M_l$, latency model. $M_p$, Power model. $Net$, the network parameters from the user. $LO$, the user's latency objective. $PO$, the user's power objective. 
\begin{algorithmic}[1]
\State $L_{opt} \gets 0$
\State $P_{opt} \gets 0$
\For{Each $Config_g \in Set_{config}$}
    \State $L_{g} \gets M_l(Net, Config_g)$
    \State $P_{g} \gets M_p(Net, Config_g)$
    \State $Update \gets False$
    \If{$L_{opt}~==~0$~and~$P_{opt}~==~0$}
        \State $Update \gets True$
    \Else
        \If{($L_{opt}~>~LO$~and~$P_{opt}~>~PO$) or ($L_{opt}~<~LO$~and~$P_{opt}~<~PO$)}
            \If{$L_{g}~<~L_{opt}$~and~$P_{g}~<~P_{opt}$}
                \State $Update \gets True$
            \EndIf
        \Else
            \If{$L_{opt}~>~LO$~and~$P_{opt}~<~PO$}
                \If{$L_{g}~<~L_{opt}$~and~$P_{opt}~<~PO$}
                    \State $Update \gets True$
                \EndIf
            \Else
                \If{$P_{g}~<~P_{opt}$ and $L_{opt}~<~LO$}
                    \State $Update \gets True$
                \EndIf
            \EndIf
        \EndIf
    \EndIf
    \If{$Update$}
        \State $L_{opt} \gets L_{g}$
        \State $P_{opt} \gets P_{g}$
        \State $Config_{opt} \gets Config_{g}$
    \EndIf
\EndFor
\State \Return $Config_{opt}$
\end{algorithmic}
\label{SelectionAlgorithm}
\end{algorithm}

The trained G and D are used by the design explorer. When performing the inference, given the network parameters and the objectives, G can output the configurations encoded by one-hot. Since ordinary one-hot encoding outputs the probabilities of each choice of each configuration, we use another number between 0 and 1 called \textbf{Probability Threshold} (such as 0.2), to allow multiple sets of generated configurations output from G, which are the candidate configuration sets shown in Figure~\ref{SystemOverview}. For each configuration, if the one-hot output of one choice excesses the probability threshold, the choice is employed. Then the candidate configuration sets are the combinations of all the employed choices of all the configurations. 
For example, assume there are only two kinds of configurations that are the PE number and the SRAM size, the configurations are denoted as (PE number, SRAM size). If the one-hot outputs of ``PE number = 4'', ``PE number = 16'', ``SRAM size = 2KB'', and ``SRAM size = 8KB'' are more than 0.2, there are four sets of configurations are candidates, which are (4, 2KB), (4, 8KB), (16, 2KB), and (16, 8KB).

\subsection{Selection Algorithm}
\label{sec:selalg}

Since normally there are thousands of the candidate configuration sets after performing the inference on G in the design explorer, the design selector, which is a program, is proposed to select the most optimized set of the configurations. 
%The algorithm of the design selector is shown in Algorithm~\ref{SelectionAlgorithm}.

In Algorithm~\ref{SelectionAlgorithm}, we keep updating two variables $L_{opt}$ and $P_{opt}$ recording the optimized latency and power objectives (Lines 1-2), respectively. For each set of the configurations $Config_g$ in the candidate configurations sets (Line 3), the corresponding actual latency $L_g$ and power $P_g$ are calculated by the design model (Lines 4-5). The algorithm maintains a variable $Update$ to indicate if $L_{opt}$ and $P_{opt}$ need to be updated (Line 6). For example, if $L_{opt}$ and $P_{opt}$ have never been updated, they need to be initialized for the first time (Lines 7-8). Otherwise, there are three scenarios to perform the updating.
When both current optimized objectives ($L_{opt}$ and $P_{opt}$) are worse/better than the user's objectives ($LO$ and $PO$), it is the first scenario. In this scenario, if the objectives ($L_g$ and $P_g$) of $Config_g$ this time are both better than $L_{opt}$ and $P_{opt}$, respectively, $L_{opt}$ and $P_{opt}$ need to be updated (Lines 10-13).
This can ensure that the optimized objectives ($L_{opt}$ and $P_{opt}$) keep changing for the better. 
In contrast, The second scenario is that the current optimized latency $L_{opt}$ is worse than the user's objective $LO$, while the current optimized power $P_{opt}$ satisfies the objective $PO$. In this case, if the latency $L_g$ of $Config_g$ this time is better than the current optimized latency $L_{opt}$, and the power $P_g$ still satisfies the user's objective $PO$, even though $P_g$ might be worse than the current optimized power $P_{opt}$, the updating is still performed (Lines 15-18). This is because for the second scenario, the most significant priority is to make all the objectives satisfied.
The third scenario (Lines 20-22), where the conditions of the latency and power are reversed, is handled similarly to the second scenario.
Finally, the optimized configurations $Config_{opt}$ that have the most optimized latency $L_{opt}$ and power $P_{opt}$ are returned.

\section{Experiments}
\label{sec:exp}

\subsection{Experimental Setup}
The experiments are conducted on Intel Xeon Silver 4210R CPU @ 2.40GHz with 64GB DDR4 memory, and NVIDIA GeForce RTX 3090. The synthesis tool is Vivado 2018.3. The GAN of the framework is deployed and trained using PyTorch~\cite{pytorch}. 

\subsubsection{Design Model}

In our experiments, convolutional neural networks (CNNs) are used as the neural networks that will be executed on the neural network accelerators to be generated, since CNN is one of the most popular neural networks. 
There are two design models of the neural network accelerators involved in our experiments, where the data-flow is output stationary, but other design models can also be applied to \thename{}. 

\textbf{im2col model:} This design model adopts the \textit{im2col} computation pattern \cite{chellapilla2006high} similar to GPU computation. As shown in Table~\ref{tab:mainmodelstr}, this model explores the limits of design space and includes configurable parameters as many as possible. The establishment of this model is based on the roofline model, which analyzes the relationship between DRAM bandwidth, SRAM bandwidth, and on-chip computation. This model is used to illustrate the effectiveness of \thename{} in high dimension design space, and can represent the general behaviors of the hardware architectures with im2col computation patterns. Therefore, this model does not correspond to a specific RTL design template in our experiments.

\textbf{DnnWeaver model:} Based on the open-source DnnWeaver code~\cite{dnnweaverrepo} of the systolic array as the design template, the DnnWeaver model is established. The model is calibrated by the simulation and synthesis results of code to ensure the output of the model is aligned with the metrics of hardware employment. 
After the configurations are found by \thename{}, they can be written as parameters in the RTL code of DnnWeaver.
Note that the number of the configurations for the DnnWeaver model is larger than that in the paper of DnnWeaver, since we extend the allowed configurations of the DnnWeaver code. This model is used to make a comparison between \thename{} and DnnWeaver.

Table~\ref{tab:mainmodelstr} shows the detailed network parameters, configurations, and objectives of the design models in our experiments. Note that the latency is the end-to-end latency, which considers 3 pipelined phases for each tile including loading data, computing, and writing back the data. Besides, the objectives need to be set as the upper bound that the user can accept. For example, when the user's latency and power objectives are set as $LO$ and $PO$, respectively, \thename{} considers that the user needs latency$\leq LO$ and power$\leq PO$.

\begin{table}[htbp]
%\scriptsize
\centering
\caption{The network parameters, the configurations, and the objectives used in our experiments. (*The configuration is not included in DnnWeaver model.)}
\label{tab:mainmodelstr}
\resizebox{0.68\textwidth}{!}{
\begin{tabular}{|c|c|}
\hline
\multirow{2}{3.0cm}{\centering Network Parameters (CNN)} & \multirow{2}{7.5cm}{Input Channel, Output Channel, Output Width, Output Height, Kernel Width, Kernel Height}\\
& \\\hline
\multirow{3}{3.0cm}{\centering{Configurations: Architecture Parameters}} & \multirow{3}{7.5cm}{PE Number, SRAM to DRAM Bandwidth*, DRAM to SRAM Bandwidth*, Input SRAM Size, Weight SRAM Size, Output SRAM Size}\\
& \\
& \\\hline
\multirow{3}{3.0cm}{\centering Configurations: Mapping Strategies} & \multirow{3}{7.5cm}{Tiling Input Channel*, Tiling Output Channel*, Tiling Output Width*, Tiling Output Height*, Tiling Kernel Width*, Tiling Kernel Height*}\\
& \\
& \\\hline
{\centering Objectives} & Latency, Power \\\hline
\end{tabular}
}
% }
%\vspace{-3ex}
\end{table}

\subsubsection{Dataset}

The dataset is generated by Dataset Generator discussed in Section~\ref{sec:tphase}, where the values of the network parameters, the architecture parameters, and the mapping strategies are evenly sampled to cover most of the scenarios. In this case, various CNN layers can be learned. Then, the objectives are calculated by the design model. For im2col model, the dataset contains 23420 data for the training and 1000 for the testing. For DnnWeaver model, the dataset contains 31250 data for the training and 1000 for the testing.
Note that \thename{} does not restrict the design model, network parameters, configurations, and objectives, so other parameters (e.g. stride, padding, etc.) can also be supported with corresponding design models. When generating the dataset, with the number of the parameters and configurations increasing, the sample rate can be reduced for the trade-off between the training time and the number of scenarios covered. 
Besides, \thename{} does not target a specific CNN architecture, and various architectures generated in the dataset will be tested.
The example parts of the dataset are shown in Tables{~\ref{tab:datasetex}} and{~\ref{tab:datasetex_dnn}}, where the latency and power are normalized by the standard deviation.

\begin{table}[htbp]
%\scriptsize
\centering
\caption{An example part of the dataset of im2col model. (The abbreviations of the columns are the combinations of the first letters of the names in Table~\ref{tab:mainmodelstr}.)}
\label{tab:datasetex}
\resizebox{1\textwidth}{!}{
\begin{tabular}{cccccccccccccccccccc}
\hline
IC & OC & OW & OH & KW & KH & PEN & SDB & DSB & ISS & WSS & OSS & TIC & TOC & TOW & TOH & TKW & TKH  & L & P \\\hline
32 & 32 & 32 & 32 & 1 & 1 & 2048 & 32 & 64 & 512 & 512 & 512 & 64 & 16 & 16 & 128 & 4 & 1  & 0.0014 & 0.1098 \\
32 & 32 & 32 & 32 & 1 & 1 & 1024 & 32 & 128 & 512 & 512 & 512 & 16 & 64 & 16 & 16 & 5 & 5 & 0.0018 & 0.0977\\
... & ... & ... & ... & ... & ... & ... & ... & ... & ... & ... & ... & ... & ... & ... & ... & ... & ... & ... & ...\\
64 & 32 & 64 & 64 & 5 & 5  & 512 & 128 & 128 & 4096 & 2048 & 4096 & 64 & 64 & 16 & 256 & 5 & 1  & 0.0551 & 2.4652\\
64 & 32 & 64 & 64 & 5 & 5  & 2048 & 128 & 512 & 4096 & 2048 & 4096 & 16 & 64 & 64 & 64 & 1 & 1  & 0.0137 & 3.9144\\
... & ... & ... & ... & ... & ... & ... & ... & ... & ... & ... & ... & ... & ... & ... & ... & ... & ... & ... & ...\\\hline

\end{tabular}
}
% }
%\vspace{-3ex}
\end{table}

\begin{table}[htbp]
%\scriptsize
\centering
\caption{An example part of the dataset of DnnWeaver model. (The abbreviations of the columns are the combinations of the first letters of the names in Table~\ref{tab:mainmodelstr}.)}
\label{tab:datasetex_dnn}
\resizebox{0.6\textwidth}{!}{
\begin{tabular}{cccccccccccc}
\hline
IC & OC & OW & OH & KW & KH & PEN & ISS & WSS & OSS & L & P \\\hline
16 & 16 & 16 & 16 & 3 & 3 & 8 & 128 & 128 & 128 & 0.0171 & 1.0237\\
16 & 16 & 16 & 16 & 3 & 3 & 16 & 128 & 128 & 128 & 0.0144 & 1.2063\\
... & ... & ... & ... & ... & ... & ... & ... & ... & ... & ... & ...\\
16 & 64 & 32 & 32 & 3 & 3 & 16 & 512 & 2048 & 512 & 0.2462 & 1.2201\\
16 & 64 & 32 & 32 & 3 & 3 & 32 & 512 & 2048 & 512 & 0.2463 & 1.2157\\
... & ... & ... & ... & ... & ... & ... & ... & ... & ... & ... & ...\\\hline

\end{tabular}
}
% }
%\vspace{-3ex}
\end{table}

\subsubsection{Hyperparameters and Training}

Table~\ref{tab:hyper} listed the hyperparameters when training the proposed GAN in \thename{}, where both G and D are multilayer perceptrons. The training phase only needs to be performed for each design template.
In our experiments, we only tune the number of the layers, the number of the neurons in each layer (which is the same in each layer, and is tuned by setting it as 512/1024/2048/4096), and the learning rate (which is tuned by setting it in the range of 1e-5 to 1e-4).
Therefore, simple hyperparameter tuning is already enough to make \thename{} have good DSE results after training.

\begin{table}[htbp]
%\scriptsize
\centering
\caption{The hyperparameters of the GAN. ($w_{critic}$ is not listed as it is separately discussed in the results; The batch sizes under both design models are 1024.)}
\label{tab:hyper}
\resizebox{1\textwidth}{!}{
\begin{tabular}{c|ccccc|ccccc}
\hline
 & \multicolumn{5}{c|}{G} & \multicolumn{5}{c}{D} \\\hline
\multirow{2}{*}{\centering Design Model} & Hidden & Neurons & Learning & Activation & \multirow{2}{*}{\centering Optimizer} & Hidden & Neurons & Learning & Activation & \multirow{2}{*}{\centering Optimizer} \\
& Layers & Per Layer & Rate & Function & & Layers & Per Layer & Rate & Function &  \\\hline
im2col & 11 & 2048 & 2e-5 & ReLU & Adam & 11 & 2048 & 2e-5 & ReLU & Adam  \\\hline
DnnWeaver& 14 & 2048 & 2.5e-5 & ReLU & Adam & 11 & 2048 & 2.5e-5 & ReLU & Adam  \\\hline
\end{tabular}
}
% }
%\vspace{-3ex}
\end{table}

%The G and D of GAN for im2col model are fully-connected neural networks, with 11 hidden layers, each with 2048 neurons for both G and D. The activation function of each layer is ReLU.

%The G and D of GAN for DnnWeaver model are fully-connected neural networks, with 14 and 11 hidden layers, each with 2048 neurons for G and D, respectively. The activation function of each layer is ReLU.

%There are 2048 test results, while the training samples in the main model and DnnWeaver model are 23420 and 31250 respectively.

%To show a contrast with the \thename{}, following methods are as baselines.

\subsubsection{Compared DSE Algorithms}

The following DSE algorithms are used for both im2col model and DnnWeaver model for comparison. %The algorithms include exhaustive searching that is used by DnnWeaver, and simulated annealing that is a famous searching algorithm.

\begin{itemize}[leftmargin=2ex]
%\item \textbf{Exhaustive searching (ES)}. Exhaustive searching is employed in many recent neural network accelerator design automation frameworks~\cite{Sharma16,Zhang18,Abdelouahab16,Sohrabizadeh21}, such as DnnWeaver. In our experiment, ES terminates once the user's objectives are satisfied, or the runtime excesses 5 minutes.
\item \textbf{Simulated annealing (SA)}. Simulated annealing is used for comparison. SA terminates once the user's objectives are satisfied, or the temperature is $3\times 10^{-8}$ as the initial one.
\item \textbf{Deep reinforcement learning (DRL)}.
For DRL, similar to ConfuciuX~\cite{Kao20}, the policy gradient is employed in our experiment, and a neural network is used as the actor network to update the underlying policy network. The states are the current network parameters and  configurations, and the actions are the modifications to the configurations. The reward is obtained when the current action is approaching the states that satisfied the objectives. When the current state already satisfies the objectives, a bonus is also added to the reward.
%gap between latency/power under current configurations and targets is set as reward in DRL. 
%, which could maximize the probability of receiving better reward is applied. 
After performing hyperparameter tuning, the relatively better results are shown in our experiments. 
%Since the tasks of DRL and GAN are different, the DRL with relatively better results has fewer parameters than GAN.
%which configuration to change and the changing direction. The gap between latency/power under present configurations and targets is set as reward in DRL. 
% To show the effectiveness of \thename{}, the number of the parameters in the DRL is also set to be the same as that in the GAN of \thename{}.
%Simulated annealing is performed around 13 million iterations, which ensures that simulated annealing has the same magnitude of the time cost as exhaustive searching.
%Around 13 million iterations ensure that SA has the same  in the same searching time compared with exhaustive searching.
\item \textbf{Large multilayer perceptron (Large MLP)}. Similar as AIRCHITECT~\cite{Samajdar21}, we also tested the results when only applying the MLP as shown in Figure~\ref{fig:train}(a). Besides, we also apply the design selector to improve the results. To show the effectiveness of \thename{}, the number of the parameters in the MLP is set to match that in the GAN, which makes the MLP much larger than the G in the GAN. The training epochs of both GAN and large MLP are also set to the same.
\end{itemize}

All the compared DSE algorithms are modified to perform DSE based on the same system-level architectures (design templates) as \thename{} for fair comparison.

\subsection{\thename{} Results Analysis}

We first analyze the effectiveness of the GAN-based DSE algorithm in \thename{}. The results are shown in Table~\ref{tab:mainresults}, where the bold numbers are the main results. When training GAN, different values of $w_{critic}$ in Algorithm~\ref{alg:gantrain} are chosen. 
Note that we allow $1\%$ of the noise when evaluating all DSE algorithms in Table~\ref{tab:mainresults}, which means if the latency/power of the DSE output is $\leq 1\%$ worse than the user's objective, the objective is still regarded as satisfied.

\begin{table*}[htbp]
%\scriptsize
\centering
\caption{The DSE results of different methods under two  design models. ($^\dagger$``Cand. Config." stands for candidate configurations. $^\ddagger$``Param." stands for parameters; The NN is the neural network used for DSE, but not the one to be executed on the accelerator.)}
\label{tab:mainresults}
\resizebox{1\textwidth}{!}{
\begin{tabular}{|cc|c||c|c|c|c|c|c|c|}
\hline
 & &  \multirow{2}{1.6cm}{\centering Method}  &  \multirow{2}{1cm}{\centering $w_{critic}$} & \multirow{2}{1.5cm}{\centering Training Time} &  \multirow{2}{1.4cm}{\centering \# of Cand. Config.$^\dagger$}  &   \multirow{2}{1.4cm}{\centering \# of NN Param.$^\ddagger$}   &   \multirow{2}{1cm}{\centering DSE Time}   &   \multirow{2}{1.5cm}{\centering \# of Sat. Results}   &   \multirow{2}{1.8cm}{\centering Improvement Ratio}  \\ 
 &    &   & &  &   &   &   &   &  \\ \hline
\multirow{8}{0.1cm}{\rotatebox[origin=c]{90}{im2col}} & \multirow{8}{*}{\rotatebox[origin=c]{90}{(Large)}} &  SA   &  - & - &  -  &   -   &  17.60s  &   786/1000   & 0.2526 \\ \cline{3-10}
%  & & DRL$^*$ &  -  &  16140s  & - &  3M  &  0.07s  &   731/1000   & 0.3137 \\ \cline{3-10}
 & &  DRL  &  -  &  133709s  & - &  93M  &  0.08s  &   687/1000   & 0.3399 \\ \cline{3-10}
 & &  Large MLP  &  - & 47267s & 10794.01 &  93M  &  1.11s  &   839/1000   & 0.3281 \\ \cline{3-10}
& &  \multirow{5}{*}{\centering GAN}  & 0 & 78806s & 732.68 &  93M  &  0.09s  &   554/1000   & 0.3286 \\ \cline{4-10}
&  &   &  \textbf{0.5}  & \textbf{116229s} &  \textbf{5360.19}  &  \textbf{93M}  &  \textbf{0.59s}  &   \textbf{944/1000}   &  \textbf{0.3579}  \\ \cline{4-10}
 & &    & 0.7& 118474s & 5059.44 &  93M  &  0.74s  &   792/1000   & 0.3405 \\ \cline{4-10}
 & &     & 1 & 116228s & 6861.21 &  93M  &  0.77s  &   893/1000   & 0.3458 \\ \cline{4-10}
 & &   & 1.2 & 118566s & 9202.87 &  93M  &  1.23s  &   909/1000   & 0.3399 \\ \hline\hline
\multirow{8}{0.1cm}{\rotatebox[origin=c]{90}{DnnWeaver Model}} & \multirow{8}{*}{\rotatebox[origin=c]{90}{(Small)}} &  SA   &- &  -   &   -   &   - &  13.60s  &   936/1000   &   0.0010 \\ \cline{3-10}
%  & & DRL  &  -  & 26406s &  -  &  11M  &   <0.01s   &   917/1000   &   0.0020 \\ \cline{3-10}
 & &  DRL &  -  & 27634s &  -  &  110M  &   <0.01s   &   948/1000   &   0.0019 \\ \cline{3-10}
 & &  Large MLP  & -&  11618s   & 26.54 &  110M  &   <0.01s   &   862/1000   &   0.0015 \\ \cline{3-10}
 & &  \multirow{5}{*}{\centering GAN}  & 0 & 173345s & 20.17 &  105M  &   <0.01s   &   847/1000   &   0.0017 \\ \cline{4-10}
 & &  & 0.5 & 36345s & 21.17 &  105M  &   <0.01s   &   938/1000   &  0.0014 \\ \cline{4-10}
 & &  &  \textbf{1.0} & \textbf{36298s}  &   \textbf{18.42}  &  \textbf{105M}  &   \textbf{<0.01s}   &   \textbf{964/1000}   &   \textbf{0.0015} \\ \cline{4-10}
 & &  & 1.2 & 36421s & 20.52 &  105M  &   <0.01s   &   945/1000   &   0.0016 \\ \cline{4-10}
 & &  & 1.5 & 25224s & 14.2 &  105M  &   <0.01s   &   938/1000   &   0.0016 \\ \hline

\end{tabular}
}
%\vspace{-2ex}
\end{table*}

Under im2col model, different $w_{critic}$ values have different effects on the GAN-based DSE algorithm. When $w_{critic}=0$, the D has no effect in the GAN, and the training scheme is the same as Figure~\ref{fig:train}(a) with MLP only. Given 1000 DSE tasks, there are only around 50\% results that satisfy the user's objectives. 
In contrast, when $w_{critic}=0.5$, the D can help the learning of G, and almost all the results successfully satisfy the objectives.
When further increasing $w_{critic}$, the results start getting slightly worse but still much better than that without D. This is because the satisfaction classification result from D contributes too much to the loss of G. It becomes hard for G to directly learn from the training set. The results show the effectiveness of \thename{} for finding the optimized configurations, and by hyperparameter tuning, \thename{} can be further improved with a proper $w_{critic}$ ($0.5$ under im2col model). 

Table~\ref{tab:mainresults} shows the average runtime of each DSE task. The GAN-based DSE algorithm with $w_{critic}=0.5$ only needs $<1s$ to complete one DSE task, which is negligible in the hardware design flow. Besides, the value of $w_{critic}$ can affect the runtime. This is mainly because of the different numbers of candidate configurations.

\begin{sloppypar}
Besides, the improvement ratio is also analyzed. \thename{} can find the configurations that outperform the user's objectives. Assume the latency and power objectives of the configurations from \thename{} are $L_{opt}$ and $P_{opt}$, respectively, and the user's objectives are $LO$ and $PO$, the improvement ratio is defined as $\sqrt{\frac{1}{2}\times ((\frac{L_{opt}-LO}{LO})^2+(\frac{P_{opt}-PO}{PO})^2)} $ when $L_{opt}\leq LO$ and $P_{opt}\leq PO$.
%\begin{center}\vspace*{-1mm}
%$\sqrt{\frac{1}{2}\times %((\frac{L_{opt}-LO}{LO})^2+(\frac{P_{opt}-PO}{PO})^2)} %$\\
%\end{center}
This can indicate how much the results of \thename{} outperform the objectives. 
When one result is not satisfied, there is at least one objective (latency or power) worse than the user's objectives, so the result is invalid and there is no improvement ratio for this result.
Shown in Table~\ref{tab:mainresults}, with a proper $w_{critic}$, the results of \thename{} are more than 35\% better than the user's objectives on average. This also proves the effectiveness of \thename{}.
\end{sloppypar}

The training time of \thename{} is more than large MLP, but it is still acceptable. Once it is trained, it can be used for any times for high quality designs given different objectives.
Meanwhile, compared with the time period needed by the hardware design, which are counted by months, the training time is already negligible. 

\begin{figure}[!h]

\centering
\includegraphics[width=0.7\columnwidth]{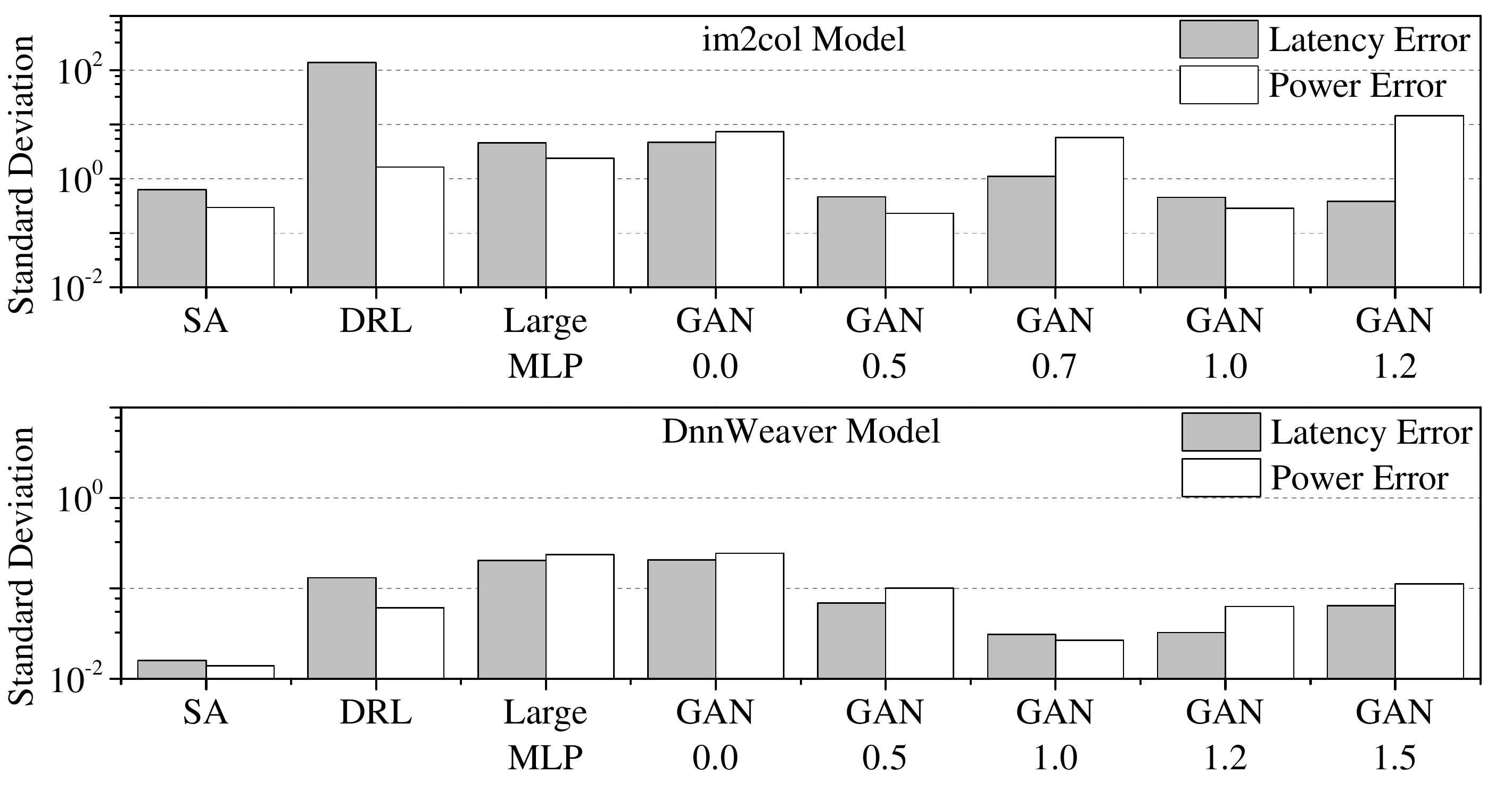}
\caption{The standard deviations of latency and power errors. (The numbers under ``GAN'' is $w_{critic}$.)}
\label{fig:restd}
\vspace{-2ex}
\end{figure}

Figure~\ref{fig:restd} shows the standard deviations of latency and power errors. The latency and power errors are defined as $\frac{L_{opt}-LO}{LO}$ and $\frac{P_{opt}-PO}{PO}$, respectively. It is indicated that for im2col model, GAN-based DSE has the least standard deviation when $w_{critic}=0.5$. This also proves that D can help the regression of the results of G, compared training G only when $w_{critic}=0$. 

We also test the effectiveness of \thename{} on DnnWeaver model, which has a much lower dimension of design space. For DnnWeaver model, the optimized design can be obtained by training the GAN with $w_{cricit}=1.0$ in Table~\ref{tab:mainresults}. If $w_{cricit}=0$ and the training of G is not affected by the D, compared with the results of $w_{critic}=1.0$, fewer results satisfy the user's objectives. Meanwhile, when $w_{critic}$ is too large, the results are also slightly worse than $w_{critic}=1.0$, but is still much better than $w_{critic}=0$.
Besides, shown in Figure~\ref{fig:restd}, when training GAN, the setup with the best results also has small enough standard deviations for latency and power errors, proving the effectiveness of the regression. Although SA has a smaller standard deviation, the number of the satisfied results and the improvement ratio are both worse than GAN. More importantly, compared with GAN, SA performs worse in all metrics under the large design model.
The average runtime under DnnWeaver model is around 0.01s, which is negligible. The improvement ratio of GAN on DnnWeaver model is not as obvious as large im2col model, but most of the results still satisfy the user's objectives.
Overall, the results under DnnWeaver model also indicate the effectiveness of the proposed DSE algorithm and the automation framework, and \thename{} has more advantages on the DSE within high dimension large design space.

\subsection{Comparison with Other DSE Algorithms}

We also compared \thename{} with other DSE algorithms including SA, DRL, and large MLP. The results are shown in Table~\ref{tab:mainresults} and Figure~\ref{fig:restd}.
For Large MLP, although the parameter size of this MLP is the same as or even more than that of the GAN in \thename{}, the large MLP performs worse compared with GAN. First, large MLP finds almost 100 fewer satisfied results under both models, compared with \thename{}. One can notice that MLP has a much larger candidate configuration set size, which implies that MLP has less confidence in the results. Next, large MLP also have a worse improvement ratio. Large MLP have almost 1x more runtime under im2col model. Finally, the standard deviation of the MLP results shown in Figure~\ref{fig:restd} is also much larger than that of \thename{}'s results.
This proves that the effectiveness of GAN-based DSE cannot be achieved by simply increasing the size of a MLP to the same level, without the guidance of D network. 
%For DRL, after performing hyperparameter tuning, the relatively better DRL results are listed in our experiments. We tests DRL with both small and large neural networks. For DnnWeaver model, DRL with larger neural networks have better results than that with smaller neural networks. For im2col model, large neural networks do not help DRL find the satisfied results, but smaller neural networks perform better. Compared with GAN-based DSE, Although DRL also has smaller runtime, it has worse results and standard deviations under both im2col and DnnWeaver model. Under DnnWeaver model, DRL has a slightly better improvement ratio, but its improvement ratio under im2col model is still worse than GAN-based DSE. 
For DRL, after performing hyperparameter tuning, the relatively better DRL results are obtained in our experiments. Although DRL also has smaller runtime, it has worse results and standard deviations under both im2col and DnnWeaver model, compared with the GAN-based DSE. Under DnnWeaver model, DRL has a slightly better improvement ratio, but its improvement ratio under im2col model is still worse than GAN-based DSE.
For SA, it performs better under the small model than the large model, but the detailed number of satisfied results, the standard deviation of the results, the runtime, and the improvement ratio are all worse than GAN-based DSE.
%For ES, it has the worst results compared with other algorithms, even with the largest runtime.
In conclusion, GAN-based DSE has overall better performance in the comparison.

\subsection{DSE Results with Different Objective Difficulties}

\begin{figure*}[!h]
\centering
\includegraphics[width=0.7\columnwidth]{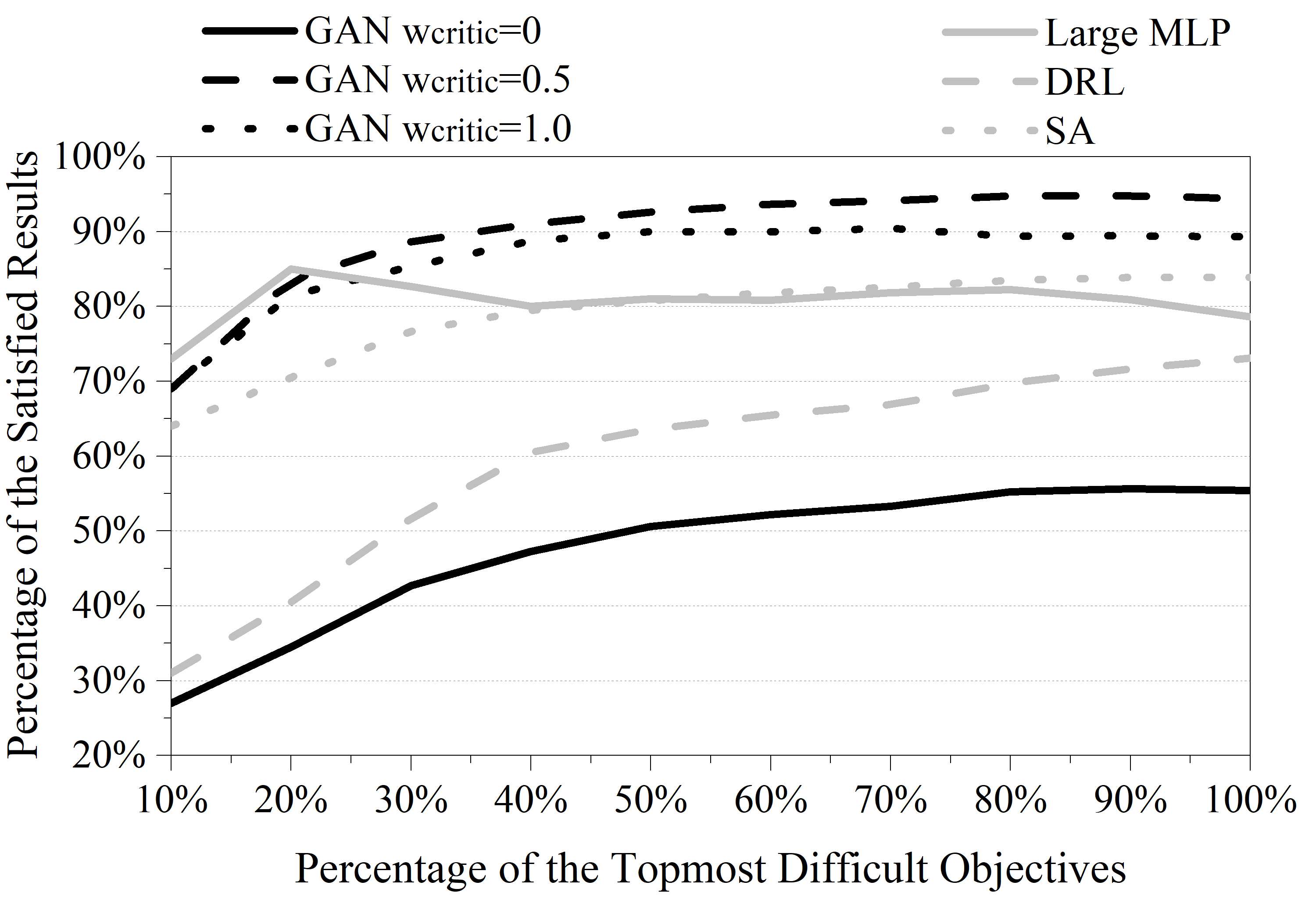}
%\caption{Latency and power improvements under different approaches in main model, and the value represents logarithm of target latency and power to generated ones}
\caption{The percentage of the satisfied results of the objectives with different difficulties under im2col model.}
\label{fig:satcurvemain}
%\vspace{-2ex}
\end{figure*}

\begin{figure*}[!h]
\centering
\includegraphics[width=0.7\columnwidth]{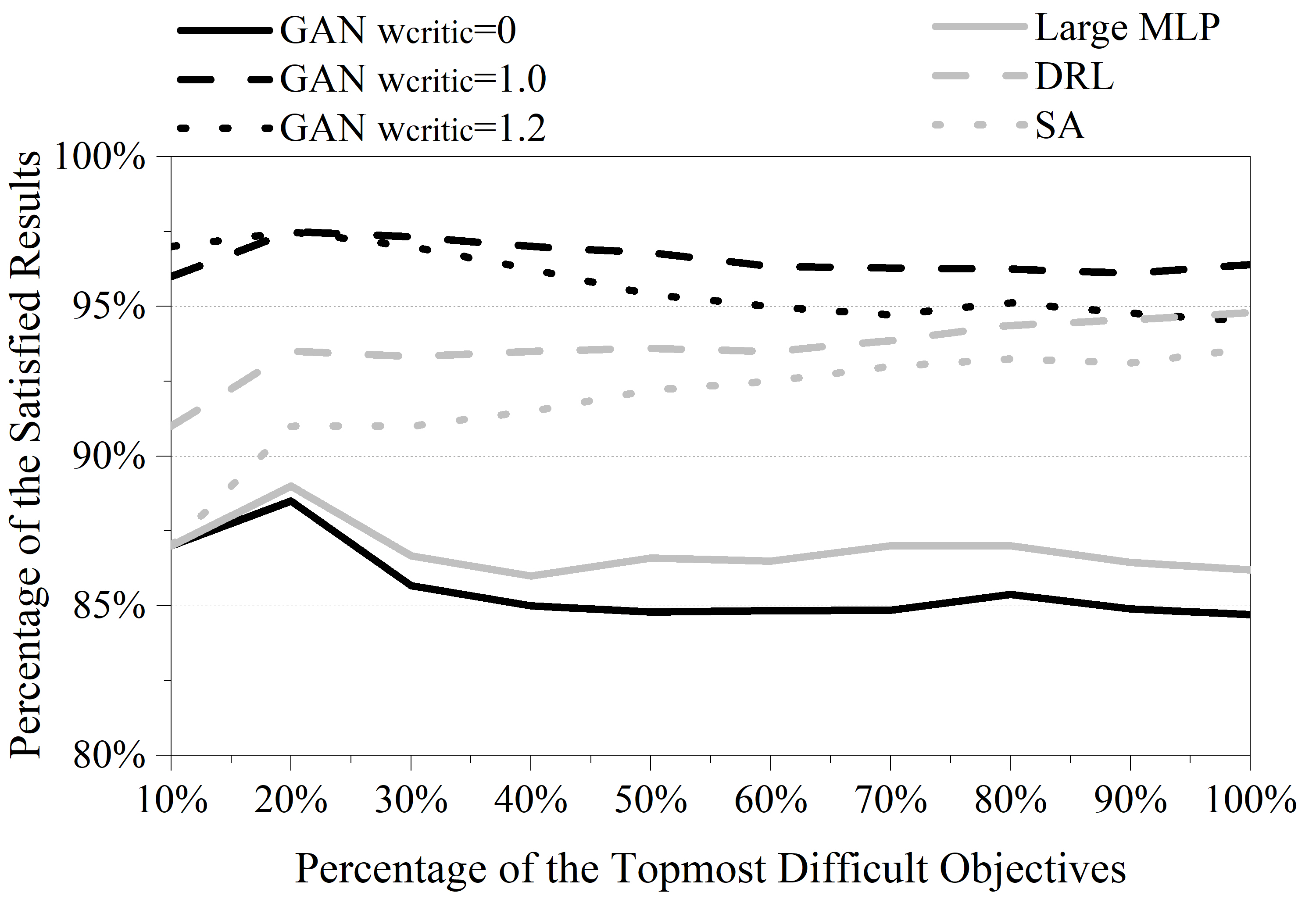}
%\caption{Latency and power improvements under different approaches in main model, and the value represents logarithm of target latency and power to generated ones}
\caption{The percentage of the satisfied results of the objectives with different difficulties under DnnWeaver model.}
\label{fig:satcurvednnw}
%\vspace{-2ex}
\end{figure*}

We also analyze the effectiveness of \thename{} given the user's objectives with different difficulties.
%and the results are shown in Figure~\ref{fig:satcurvemain} and Figure~\ref{fig:satcurvednnw}. 
First, the difficulty is quantified as follows.
There are multiple Pareto frontiers in the dataset. (For a sample $s$, if there is no other sample with one objective better than $s$ and other objectives not worse than $s$, $s$ is a Pareto frontier). 
%Each point $(LO, PO)$ of the user's latency and power objectives has a Pareto frontier with latency and power $(L_{pareto}, P_{pareto})$ that is closest to $(LO, PO)$ among all Pareto frontiers.
%For the user's latency and power objectives $(LO, PO)$, the distance between it and a Pareto frontier with objectives $(L_{pareto}, P_{pareto})$ is the Euclidean distance between points $(LO, PO)$ and $(L_{pareto}, P_{pareto})$.
The difficulty of the user's latency and power objectives $(LO, PO)$ is quantified by using the Euclidean distance between points $(LO, PO)$ and $(L_{pareto}, P_{pareto})$, which is the closest Pareto frontier to $(LO, PO)$. Besides, the Euclidean distance is then normalized by the module of $(L_{pareto}, P_{pareto})$. 
%, where the distance is also normalized by the module of the objective vector of the closest Pareto frontier. 
%Assume the user's latency and power objectives are $LO$ and $PO$, respectively, the distance to Pareto frontier is defined as the Euclidean distance $\sqrt{(LO-L_{pareto-i})^2+(PO-P_{pareto-i})^2}$, where ($L_{pareto-i}$, $P_{pareto-i}$) is the closest Pareto frontier to ($LO$, $PO$).
%Besides, we also uses the module of the objective vector of the closest Pareto frontier to normalize the Euclidean distance as $\frac{\sqrt{(LO-L_{pareto-i})^2+(PO-P_{pareto-i})^2}}{\sqrt{L_{pareto-i}^2+P_{pareto-i}^2}}$.
Then, user's objectives with smaller distances to the closest Pareto frontiers have higher difficulties to be satisfied.
Figure~\ref{fig:satcurvemain} and Figure~\ref{fig:satcurvednnw} indicate the percentage of the satisfied results (y-asix) given the topmost n\% difficult user's objectives (x-asix). One can easily figure out that most DSE algorithms have better results with more objectives that are easier to be satisfied. 
Under im2col model (Figure~\ref{fig:satcurvemain}), when selecting the topmost 10\% difficult objectives (including the Pareto frontiers), only GAN ($w_{critic}=0.5$ and $1.0$), Large MLP, and SA has relatively better results. 
However, with the growing size of the objectives, GAN ($w_{critic}=0.5$ and $1.0$) has a better ratio of the satisfied results, while the curves of other approaches fluctuate and become worse than GAN. Since GAN with $w_{critic}=0$ has no D, it is not a real GAN structure and the results are worse than those with D involved. Note that with objective difficulty decreasing, Large MLP has even worse results compared with difficult objectives. This might be because it strictly learns the architectures corresponding to the objectives in the training set, and misses the opportunities to learn better architectures, which has been discussed in Section~\ref{sec:motivation}. This is even worse for easy objectives since there are more such opportunities. 
Similar to im2col model, under DnnWeaver model (Figure~\ref{fig:satcurvednnw}), the results also prove the effectiveness of GAN, even with difficult objectives. In contrast, other approaches perform worse when the objectives are difficult to be satisfied.

\subsection{Results Distribution Analysis}

\begin{figure*}[!h]
\centering
\includegraphics[width=1\columnwidth]{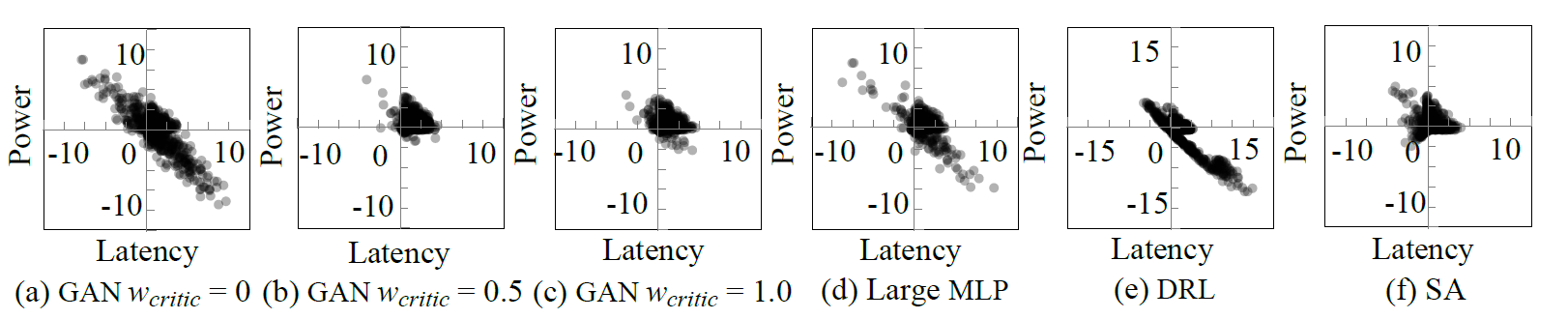}
%\caption{Latency and power improvements under different approaches in main model, and the value represents logarithm of target latency and power to generated ones}
\caption{The latency and power improvement results (dots) of the different DSE algorithms based on im2col model. A dot with non-negative coordinate of an axis represents the satisfaction of the corresponding objective.}
\label{fig:scatter_main}
\vspace{-2ex}
\end{figure*}

\begin{figure*}[!h]
\centering
\includegraphics[width=1\columnwidth]{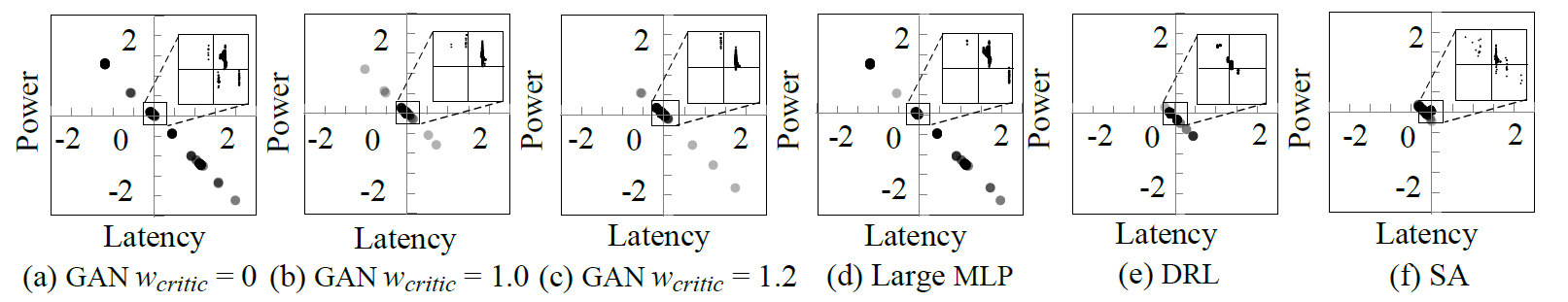}
%\caption{Latency and power improvements under different approaches in main model, and the value represents logarithm of target latency and power to generated ones}
\caption{The latency and power improvement results (dots) of the different DSE algorithms based on DnnWeaver model. A dot with non-negative coordinate of an axis represents the satisfaction of the corresponding objective.}
\label{fig:scatter_dnnw}
\vspace{-2ex}
\end{figure*}

In this subsection, we analyze the result distribution of different DSE algorithms, which is shown in Figure~\ref{fig:scatter_main} and Figure~\ref{fig:scatter_dnnw}.
Each dot is a result of the DSE, which indicates the improvements of the latency/power of the DSE output configurations compared with the user's objectives. 
%Note that the improvements represent the latency and power reduction.
The x-axis and y-axis represent $log_2(\frac{LO}{L_{opt}})$ and $log_2(\frac{PO}{P_{opt}})$, respectively.
%, where $LO$/$PO$ is the user's latency/power objective, and $L_{opt}$/$P_{opt}$ is the optimized latency/power of the output configurations of the DSE algorithm. 
%Therefore, if the x/y coordinate value of a dot is $\geq 0$, it represents the latency/power of the DSE output configurations that satisfies (or is even better than) the user's objective. 
For Figure~\ref{fig:scatter_main}(a), under im2col model, when $w_{critic}=0$, the D has no effect in the GAN. One can easily figure out that there are lots of results in the second (or fourth) quadrant, which represent they fail to satisfy the user's latency (or power) objective. 
In contrast, when $w_{critic}=0.5$ in Figure~\ref{fig:scatter_main}(b), the D can help the learning of G, and the results are well regressed into the first quadrant.
However, with the increasing of $w_{critic}$ to $1.0$ in Figure~\ref{fig:scatter_main}(c), the results are getting slightly worse. 
%This is because the satisfaction classification result from D contributes too much to the loss of G. 
This is because it becomes harder for G to directly learn from the training set than from D.

For Figure{~\ref{fig:scatter_dnnw}}, as most results are located near the origin (0,0), this part is zoomed, shown inside the rectangle at the top-right of each subfigure.
For Figure~\ref{fig:scatter_dnnw}(a)-(c), since the design space has a much smaller dimension, the results distribution under DnnWeaver model is different from that under im2col model, but more results are still in the first quadrant with the effect of D than that without D ($w_{critic}=0$), shown in the zoomed rectangle.
The best $w_{critic}$ can be obtained from hyperparameter tuning.
%Therefore, a proper $w_{critic}$ (which is $0.5$ in this design model) is needed to train the GAN. 

For im2col model, compared with Figure~\ref{fig:scatter_main}(a), large MLP in Figure~\ref{fig:scatter_main}(d) has similar distribution, but there are more dots successfully regress to the first quadrant due to larger number of parameters. However, for large MLP, there are still obvious distributions out of the first quadrant, which make the results worse than GAN-based DSE in Figure~\ref{fig:scatter_main}(b) and (c). This also proves the effectiveness of D even with fewer neural network parameters. For other previous approaches including DRL and SA, they both have worse distributions compared with GAN-based DSE since there are more dots that fail to fall in the first quadrant.
For DnnWeaver model, the related distributions are also similar to im2col model. In Figure~\ref{fig:scatter_dnnw}(d) and (e), the density of the results in the second and fourth quadrants are obviously higher than that in Figure~\ref{fig:scatter_dnnw}(b), which indicates fewer results are satisfied when using large MLP and DRL. For SA in Figure~\ref{fig:scatter_dnnw}(f), although the results are concentrated to the middle, there are still fewer results in the first quadrant compared with GAN. The distribution comparison also shows that GAN can help \thename{} find better results.

% \begin{figure}[!h]
% \centering
% \includegraphics[width=0.9\columnwidth]{mainmodel_loss_scatter.pdf}
% %\caption{Latency and power improvements under different approaches in main model, and the value represents logarithm of target latency and power to generated ones}
% \caption{The latency and power improvement results (dots) of the GAN-based DSE algorithm based on im2col model. A dot with non-negative coordinate of an axis represents the satisfaction of the corresponding objective.}
% \label{fig:scatter_loss_main}
% \end{figure}

\subsection{Training Losses}

\begin{figure}[!h]
\vspace{-1ex}
\centering
\includegraphics[width=0.7\columnwidth]{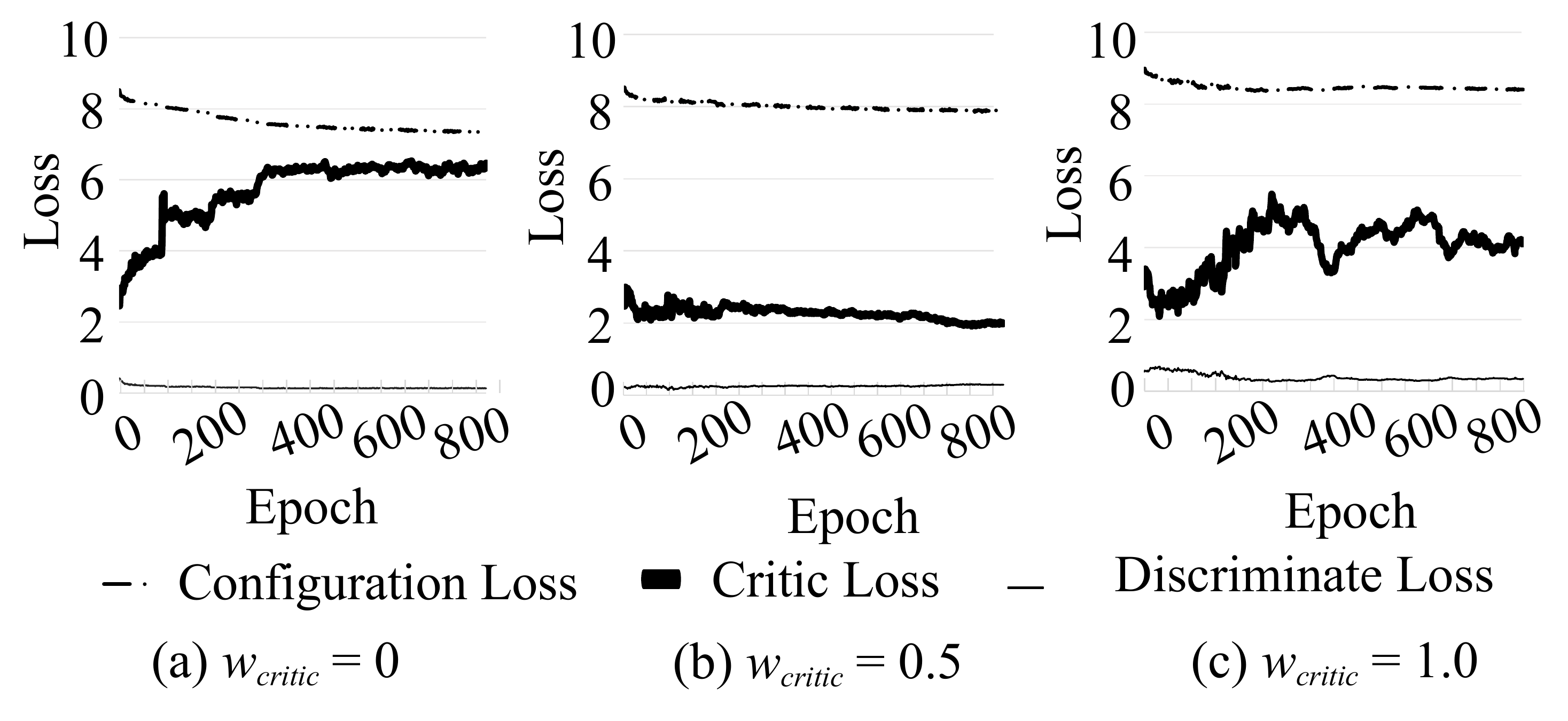}
\caption{The training losses of the GAN under im2col model.}
\label{fig:curve_loss_main}
\vspace{-2ex}
\end{figure}

\begin{figure}[!h]
\vspace{-1ex}
\centering
\includegraphics[width=0.7\columnwidth]{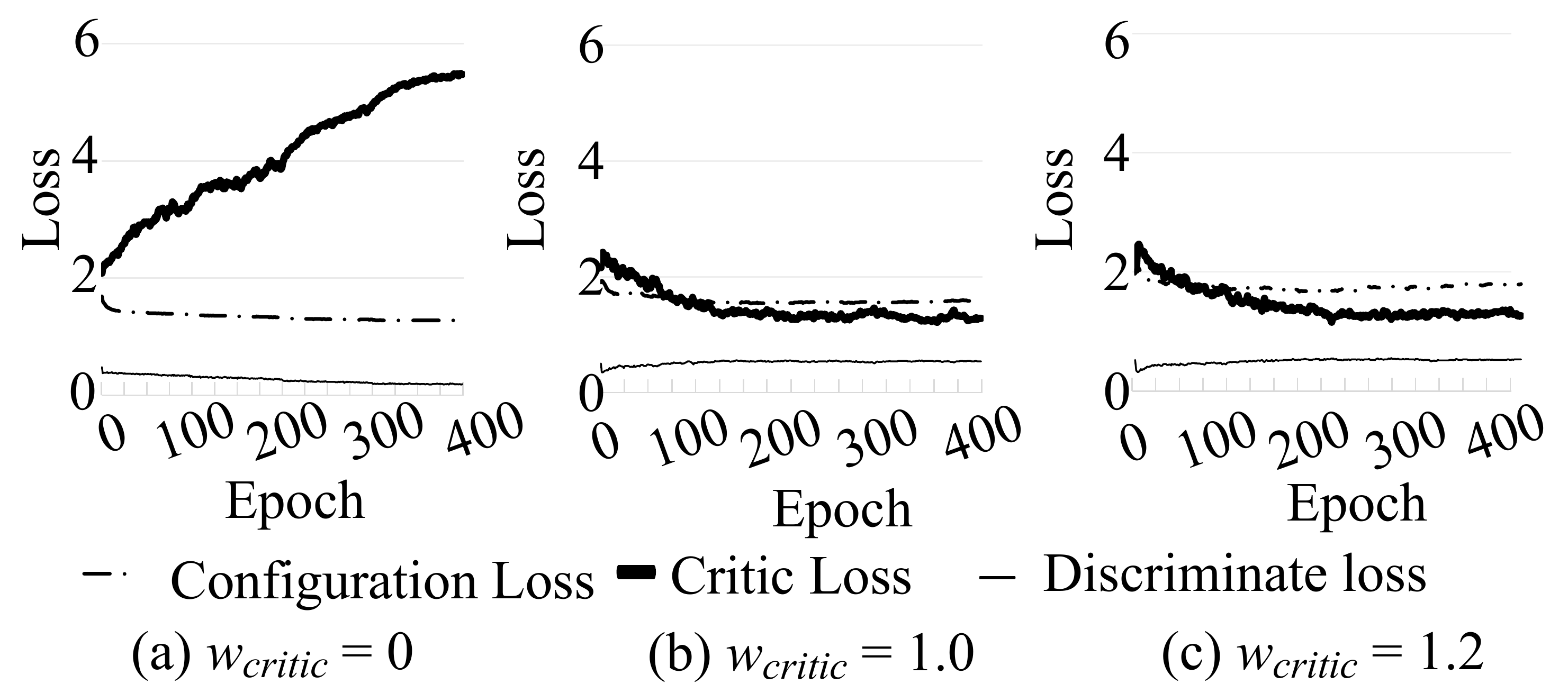}
\caption{The training losses of the GAN under DnnWeaver model.}
\label{fig:curve_loss_dnnw}
\vspace{-2ex}
\end{figure}

More training insights can be analyzed by the losses recorded through the training, which are shown in Figure~\ref{fig:curve_loss_main} and Figure~\ref{fig:curve_loss_dnnw}. Under im2col model, to analyze the G's output behaviors, when $w_{critic}=0$, the D is stilled trained, but the output of D never affect the training of G. In this case shown in Figure~\ref{fig:curve_loss_main}(a), the critic loss of G keeps increasing, which indicates that D believes some sets of the configurations output from G cannot satisfy the user's objectives. When $w_{critic}=0.5$ shown in Figure~\ref{fig:curve_loss_main}(b), since the classification result of D is used to train G, the critic loss of G is much lower than that in Figure~\ref{fig:curve_loss_main}(a). At the same time, all the losses become regressed, indicating the balance of G and D in the GAN. When $w_{critic}=1.0$ shown in Figure~\ref{fig:curve_loss_main}(c), the critic loss cannot quickly regress as the classification results of D contribute too much and the G cannot easily learn from the training set. 
Similar conclusions can also be made from the training losses under DnnWeaver model shown in Figure~\ref{fig:curve_loss_dnnw}.
The results in Figure~\ref{fig:curve_loss_main} and Figure~\ref{fig:curve_loss_dnnw} show that different $w_{critic}$ has different impacts on the training of the proposed DSE algorithm.

\section{Conclusion}
\label{sec:conclusion}

In conclusion, this work proposed a neural network accelerator design automation framework \thename{}. With the neural network accelerator template that can be configured, given the user's objectives and the architecture of the neural networks that need to be executed on the neural network accelerator, \thename{} is able to effectively search the optimized configurations of the neural network accelerator that can satisfy the objectives. A novel DSE algorithm, which is based on GAN, is proposed and applied in \thename{}. 
Compared with the previous work including using multilayer perceptron and deep reinforcement learning, the GAN-based DSE algorithm can effectively find more optimized designs within negligible time in the high dimension large design space. The experiments proved the effectiveness of \thename{}.

%%
%% The acknowledgments section is defined using the "acks" environment
%% (and NOT an unnumbered section). This ensures the proper
%% identification of the section in the article metadata, and the
%% consistent spelling of the heading.
\begin{acks}
This work was partially supported by 
National Key R\&D Program of China (Grant No.\\  2018YFE0126300),
National Natural Science Foundation of China (Grant No. 62204111, 62034007 and 62141404),
Zhejiang Provincial Key R\&D program (Grant No. 2020C01052), 
and Shuangchuang Program of Jiangsu Province (Grant No. JSSCBS20210003).
\end{acks}

%%
%% The next two lines define the bibliography style to be used, and
%% the bibliography file.
\bibliographystyle{ACM-Reference-Format}
\bibliography{references}

%%% -*-BibTeX-*-
%%% Do NOT edit. File created by BibTeX with style
%%% ACM-Reference-Format-Journals [18-Jan-2012].

\begin{thebibliography}{24}

%%% ====================================================================
%%% NOTE TO THE USER: you can override these defaults by providing
%%% customized versions of any of these macros before the \bibliography
%%% command.  Each of them MUST provide its own final punctuation,
%%% except for \shownote{}, \showDOI{}, and \showURL{}.  The latter two
%%% do not use final punctuation, in order to avoid confusing it with
%%% the Web address.
%%%
%%% To suppress output of a particular field, define its macro to expand
%%% to an empty string, or better, \unskip, like this:
%%%
%%% \newcommand{\showDOI}[1]{\unskip}   % LaTeX syntax
%%%
%%% \def \showDOI #1{\unskip}           % plain TeX syntax
%%%
%%% ====================================================================

\ifx \showCODEN    \undefined \def \showCODEN     #1{\unskip}     \fi
\ifx \showDOI      \undefined \def \showDOI       #1{#1}\fi
\ifx \showISBNx    \undefined \def \showISBNx     #1{\unskip}     \fi
\ifx \showISBNxiii \undefined \def \showISBNxiii  #1{\unskip}     \fi
\ifx \showISSN     \undefined \def \showISSN      #1{\unskip}     \fi
\ifx \showLCCN     \undefined \def \showLCCN      #1{\unskip}     \fi
\ifx \shownote     \undefined \def \shownote      #1{#1}          \fi
\ifx \showarticletitle \undefined \def \showarticletitle #1{#1}   \fi
\ifx \showURL      \undefined \def \showURL       {\relax}        \fi
% The following commands are used for tagged output and should be
% invisible to TeX
\providecommand\bibfield[2]{#2}
\providecommand\bibinfo[2]{#2}
\providecommand\natexlab[1]{#1}
\providecommand\showeprint[2][]{arXiv:#2}

\bibitem[\protect\citeauthoryear{Abdelouahab, Bourrasset, Pelcat, Berry,
  Quinton, and Serot}{Abdelouahab et~al\mbox{.}}{2016}]%
        {Abdelouahab16}
\bibfield{author}{\bibinfo{person}{Kamel Abdelouahab},
  \bibinfo{person}{C\'{e}dric Bourrasset}, \bibinfo{person}{Maxime Pelcat},
  \bibinfo{person}{Fran\c{c}ois Berry}, \bibinfo{person}{Jean-Charles Quinton},
  {and} \bibinfo{person}{Jocelyn Serot}.} \bibinfo{year}{2016}\natexlab{}.
\newblock \showarticletitle{{A Holistic Approach for Optimizing DSP Block
  Utilization of a CNN Implementation on FPGA}}.
\newblock \bibinfo{journal}{\emph{ACM International Conference on Distributed
  Smart Camera}} (\bibinfo{year}{2016}), \bibinfo{pages}{69–75}.
\newblock


\bibitem[\protect\citeauthoryear{Chellapilla, Puri, and Simard}{Chellapilla
  et~al\mbox{.}}{2006}]%
        {chellapilla2006high}
\bibfield{author}{\bibinfo{person}{Kumar Chellapilla}, \bibinfo{person}{Sidd
  Puri}, {and} \bibinfo{person}{Patrice Simard}.}
  \bibinfo{year}{2006}\natexlab{}.
\newblock \showarticletitle{{High Performance Convolutional Neural Networks for
  Document Processing}}.
\newblock \bibinfo{journal}{\emph{International Workshop on Frontiers in
  Handwriting Recognition}} (\bibinfo{year}{2006}), \bibinfo{pages}{1--7}.
\newblock


\bibitem[\protect\citeauthoryear{Chen, Krishna, Emer, and Sze}{Chen
  et~al\mbox{.}}{2017}]%
        {Chen17}
\bibfield{author}{\bibinfo{person}{{Yu-Hsin} Chen}, \bibinfo{person}{Tushar
  Krishna}, \bibinfo{person}{Joel~S. Emer}, {and} \bibinfo{person}{Vivienne
  Sze}.} \bibinfo{year}{2017}\natexlab{}.
\newblock \showarticletitle{{Eyeriss: An Energy-Efficient Reconfigurable
  Accelerator for Deep Convolutional Neural Networks}}.
\newblock \bibinfo{journal}{\emph{IEEE Journal of Solid-State Circuits}}
  \bibinfo{volume}{52}, \bibinfo{number}{1} (\bibinfo{year}{2017}),
  \bibinfo{pages}{127--138}.
\newblock


\bibitem[\protect\citeauthoryear{Chetlur, Woolley, Vandermersch, Cohen, Tran,
  Catanzaro, and Shelhamer}{Chetlur et~al\mbox{.}}{2014}]%
        {ChetlurWVCTCS14}
\bibfield{author}{\bibinfo{person}{Sharan Chetlur}, \bibinfo{person}{Cliff
  Woolley}, \bibinfo{person}{Philippe Vandermersch}, \bibinfo{person}{Jonathan
  Cohen}, \bibinfo{person}{John Tran}, \bibinfo{person}{Bryan Catanzaro}, {and}
  \bibinfo{person}{Evan Shelhamer}.} \bibinfo{year}{2014}\natexlab{}.
\newblock \showarticletitle{{cuDNN: Efficient Primitives for Deep Learning}}.
\newblock \bibinfo{journal}{\emph{arXiv Preprint}} (\bibinfo{year}{2014}),
  \bibinfo{pages}{1--9}.
\newblock


\bibitem[\protect\citeauthoryear{{DnnWeaver v2.0}}{{DnnWeaver v2.0}}{2016}]%
        {dnnweaverrepo}
\bibfield{author}{\bibinfo{person}{{DnnWeaver v2.0}}.}
  \bibinfo{year}{2016}\natexlab{}.
\newblock \bibinfo{howpublished}{\url{http://dnnweaver.org/}}.
\newblock


\bibitem[\protect\citeauthoryear{Genc, Haj-Ali, Iyer, Amid, Mao, Wright,
  Schmidt, Zhao, Ou, Banister, Shao, Nikolic, Stoica, and Asanovic}{Genc
  et~al\mbox{.}}{2019}]%
        {Genc19}
\bibfield{author}{\bibinfo{person}{Hasan Genc}, \bibinfo{person}{Ameer
  Haj-Ali}, \bibinfo{person}{Vighnesh Iyer}, \bibinfo{person}{Alon Amid},
  \bibinfo{person}{Howard Mao}, \bibinfo{person}{John Wright},
  \bibinfo{person}{Colin Schmidt}, \bibinfo{person}{Jerry Zhao},
  \bibinfo{person}{Albert Ou}, \bibinfo{person}{Max Banister},
  \bibinfo{person}{Yakun~Sophia Shao}, \bibinfo{person}{Borivoje Nikolic},
  \bibinfo{person}{Ion Stoica}, {and} \bibinfo{person}{Krste Asanovic}.}
  \bibinfo{year}{2019}\natexlab{}.
\newblock \showarticletitle{{Gemmini: An Agile Systolic Array Generator
  Enabling Systematic Evaluations of Deep-Learning Architectures}}.
\newblock \bibinfo{journal}{\emph{arXiv preprint}} (\bibinfo{year}{2019}),
  \bibinfo{pages}{1--6}.
\newblock


\bibitem[\protect\citeauthoryear{Goodfellow, Pouget-Abadie, Mirza, Xu,
  Warde-Farley, Ozair, Courville, and Bengio}{Goodfellow et~al\mbox{.}}{2020}]%
        {goodfellow14}
\bibfield{author}{\bibinfo{person}{Ian Goodfellow}, \bibinfo{person}{Jean
  Pouget-Abadie}, \bibinfo{person}{Mehdi Mirza}, \bibinfo{person}{Bing Xu},
  \bibinfo{person}{David Warde-Farley}, \bibinfo{person}{Sherjil Ozair},
  \bibinfo{person}{Aaron Courville}, {and} \bibinfo{person}{Yoshua Bengio}.}
  \bibinfo{year}{2020}\natexlab{}.
\newblock \showarticletitle{{Generative Adversarial Networks}}.
\newblock \bibinfo{journal}{\emph{Commun. ACM}} \bibinfo{volume}{63},
  \bibinfo{number}{11} (\bibinfo{year}{2020}), \bibinfo{pages}{139–144}.
\newblock


\bibitem[\protect\citeauthoryear{Jia, Shelhamer, Donahue, Karayev, Long,
  Girshick, Guadarrama, and Darrell}{Jia et~al\mbox{.}}{2014}]%
        {caffe}
\bibfield{author}{\bibinfo{person}{Yangqing Jia}, \bibinfo{person}{Evan
  Shelhamer}, \bibinfo{person}{Jeff Donahue}, \bibinfo{person}{Sergey Karayev},
  \bibinfo{person}{Jonathan Long}, \bibinfo{person}{Ross Girshick},
  \bibinfo{person}{Sergio Guadarrama}, {and} \bibinfo{person}{Trevor Darrell}.}
  \bibinfo{year}{2014}\natexlab{}.
\newblock \showarticletitle{{Caffe: Convolutional Architecture for Fast Feature
  Embedding}}.
\newblock \bibinfo{journal}{\emph{ACM International Conference on Multimedia}}
  (\bibinfo{year}{2014}), \bibinfo{pages}{675--678}.
\newblock


\bibitem[\protect\citeauthoryear{Kao, Jeong, and Krishna}{Kao
  et~al\mbox{.}}{2020}]%
        {Kao20}
\bibfield{author}{\bibinfo{person}{Sheng-Chun Kao}, \bibinfo{person}{Geonhwa
  Jeong}, {and} \bibinfo{person}{Tushar Krishna}.}
  \bibinfo{year}{2020}\natexlab{}.
\newblock \showarticletitle{{ConfuciuX: Autonomous Hardware Resource Assignment
  for DNN Accelerators using Reinforcement Learning}}.
\newblock \bibinfo{journal}{\emph{IEEE/ACM International Symposium on
  Microarchitecture}} (\bibinfo{year}{2020}), \bibinfo{pages}{622--636}.
\newblock


\bibitem[\protect\citeauthoryear{Lin, Yang, and Han}{Lin et~al\mbox{.}}{2021}]%
        {Lin21}
\bibfield{author}{\bibinfo{person}{Yujun Lin}, \bibinfo{person}{Mengtian Yang},
  {and} \bibinfo{person}{Song Han}.} \bibinfo{year}{2021}\natexlab{}.
\newblock \showarticletitle{{NAAS: Neural Accelerator Architecture Search}}.
\newblock \bibinfo{journal}{\emph{ACM/IEEE Design Automation Conference}}
  (\bibinfo{year}{2021}), \bibinfo{pages}{1051--1056}.
\newblock


\bibitem[\protect\citeauthoryear{Luo, Liu, Li, Wang, Zhang, Chen, Xu, Temam,
  and Chen}{Luo et~al\mbox{.}}{2017}]%
        {Luo17}
\bibfield{author}{\bibinfo{person}{Tao Luo}, \bibinfo{person}{Shaoli Liu},
  \bibinfo{person}{Ling Li}, \bibinfo{person}{Yuqing Wang},
  \bibinfo{person}{Shijin Zhang}, \bibinfo{person}{Tianshi Chen},
  \bibinfo{person}{Zhiwei Xu}, \bibinfo{person}{Olivier Temam}, {and}
  \bibinfo{person}{Yunji Chen}.} \bibinfo{year}{2017}\natexlab{}.
\newblock \showarticletitle{{DaDianNao: A Neural Network Supercomputer}}.
\newblock \bibinfo{journal}{\emph{IEEE Trans. Comput.}} \bibinfo{volume}{66},
  \bibinfo{number}{1} (\bibinfo{year}{2017}), \bibinfo{pages}{73--88}.
\newblock


\bibitem[\protect\citeauthoryear{Mirza and Osindero}{Mirza and
  Osindero}{2014}]%
        {Mirza14}
\bibfield{author}{\bibinfo{person}{Mehdi Mirza} {and} \bibinfo{person}{Simon
  Osindero}.} \bibinfo{year}{2014}\natexlab{}.
\newblock \showarticletitle{{Conditional Generative Adversarial Nets}}.
\newblock \bibinfo{journal}{\emph{arXiv preprint}} (\bibinfo{year}{2014}),
  \bibinfo{pages}{1--7}.
\newblock


\bibitem[\protect\citeauthoryear{{NVDLA}}{{NVDLA}}{2018}]%
        {NVDLA}
\bibfield{author}{\bibinfo{person}{{NVDLA}}.} \bibinfo{year}{2018}\natexlab{}.
\newblock \bibinfo{howpublished}{\url{http://nvdla.org/}}.
\newblock


\bibitem[\protect\citeauthoryear{Paszke, Gross, Massa, Lerer, Bradbury, Chanan,
  Killeen, Lin, Gimelshein, Antiga, Desmaison, Kopf, Yang, DeVito, Raison,
  Tejani, Chilamkurthy, Steiner, Fang, Bai, and Chintala}{Paszke
  et~al\mbox{.}}{2019}]%
        {pytorch}
\bibfield{author}{\bibinfo{person}{Adam Paszke}, \bibinfo{person}{Sam Gross},
  \bibinfo{person}{Francisco Massa}, \bibinfo{person}{Adam Lerer},
  \bibinfo{person}{James Bradbury}, \bibinfo{person}{Gregory Chanan},
  \bibinfo{person}{Trevor Killeen}, \bibinfo{person}{Zeming Lin},
  \bibinfo{person}{Natalia Gimelshein}, \bibinfo{person}{Luca Antiga},
  \bibinfo{person}{Alban Desmaison}, \bibinfo{person}{Andreas Kopf},
  \bibinfo{person}{Edward Yang}, \bibinfo{person}{Zachary DeVito},
  \bibinfo{person}{Martin Raison}, \bibinfo{person}{Alykhan Tejani},
  \bibinfo{person}{Sasank Chilamkurthy}, \bibinfo{person}{Benoit Steiner},
  \bibinfo{person}{Lu Fang}, \bibinfo{person}{Junjie Bai}, {and}
  \bibinfo{person}{Soumith Chintala}.} \bibinfo{year}{2019}\natexlab{}.
\newblock \showarticletitle{{PyTorch: An Imperative Style, High-Performance
  Deep Learning Library}}.
\newblock \bibinfo{journal}{\emph{Advances in Neural Information Processing
  Systems}}  \bibinfo{volume}{32} (\bibinfo{year}{2019}),
  \bibinfo{pages}{1--12}.
\newblock


\bibitem[\protect\citeauthoryear{Samajdar, Joseph, Denton, and
  Krishna}{Samajdar et~al\mbox{.}}{2021}]%
        {Samajdar21}
\bibfield{author}{\bibinfo{person}{Ananda Samajdar},
  \bibinfo{person}{Jan~Moritz Joseph}, \bibinfo{person}{Matthew Denton}, {and}
  \bibinfo{person}{Tushar Krishna}.} \bibinfo{year}{2021}\natexlab{}.
\newblock \showarticletitle{{AIRCHITECT: Learning Custom Architecture Design
  and Mapping Space}}.
\newblock \bibinfo{journal}{\emph{arXiv preprint}} (\bibinfo{year}{2021}),
  \bibinfo{pages}{1--12}.
\newblock


\bibitem[\protect\citeauthoryear{Schafer and Wang}{Schafer and Wang}{2020}]%
        {Schafer20}
\bibfield{author}{\bibinfo{person}{Benjamin~Carrion Schafer} {and}
  \bibinfo{person}{Zi Wang}.} \bibinfo{year}{2020}\natexlab{}.
\newblock \showarticletitle{{High-Level Synthesis Design Space Exploration:
  Past, Present, and Future}}.
\newblock \bibinfo{journal}{\emph{{IEEE Transactions on Computer-Aided Design
  of Integrated Circuits and Systems}}} \bibinfo{volume}{39},
  \bibinfo{number}{10} (\bibinfo{year}{2020}), \bibinfo{pages}{2628--2639}.
\newblock


\bibitem[\protect\citeauthoryear{Sharma, Park, Mahajan, Amaro, Kim, Shao,
  Mishra, and Esmaeilzadeh}{Sharma et~al\mbox{.}}{2016}]%
        {Sharma16}
\bibfield{author}{\bibinfo{person}{Hardik Sharma}, \bibinfo{person}{Jongse
  Park}, \bibinfo{person}{Divya Mahajan}, \bibinfo{person}{Emmanuel Amaro},
  \bibinfo{person}{Joon~Kyung Kim}, \bibinfo{person}{Chenkai Shao},
  \bibinfo{person}{Asit Mishra}, {and} \bibinfo{person}{Hadi Esmaeilzadeh}.}
  \bibinfo{year}{2016}\natexlab{}.
\newblock \showarticletitle{{From High-Level Deep Neural Models to FPGAs}}.
\newblock \bibinfo{journal}{\emph{IEEE/ACM International Symposium on
  Microarchitecture}} (\bibinfo{year}{2016}), \bibinfo{pages}{1--12}.
\newblock


\bibitem[\protect\citeauthoryear{Sohrabizadeh, Bai, Sun, and Cong}{Sohrabizadeh
  et~al\mbox{.}}{2021}]%
        {Sohrabizadeh21}
\bibfield{author}{\bibinfo{person}{Atefeh Sohrabizadeh},
  \bibinfo{person}{Yunsheng Bai}, \bibinfo{person}{Yizhou Sun}, {and}
  \bibinfo{person}{Jason Cong}.} \bibinfo{year}{2021}\natexlab{}.
\newblock \showarticletitle{{GNN-DSE: Automated Accelerator Optimization Aided
  by Graph Neural Networks}}.
\newblock \bibinfo{journal}{\emph{arXiv Preprint}} (\bibinfo{year}{2021}),
  \bibinfo{pages}{1--12}.
\newblock


\bibitem[\protect\citeauthoryear{Sohrabizadeh, Yu, Gao, and Cong}{Sohrabizadeh
  et~al\mbox{.}}{2022}]%
        {Sohrabizadeh20}
\bibfield{author}{\bibinfo{person}{Atefeh Sohrabizadeh},
  \bibinfo{person}{Cody~Hao Yu}, \bibinfo{person}{Min Gao}, {and}
  \bibinfo{person}{Jason Cong}.} \bibinfo{year}{2022}\natexlab{}.
\newblock \showarticletitle{{AutoDSE: Enabling Software Programmers to Design
  Efficient {FPGA} Accelerators}}.
\newblock \bibinfo{journal}{\emph{ACM Transactions on Design Automation of
  Electronic Systems}} \bibinfo{volume}{27}, \bibinfo{number}{4}
  (\bibinfo{year}{2022}), \bibinfo{pages}{1--27}.
\newblock


\bibitem[\protect\citeauthoryear{Xu, Zhang, Hao, Zhao, Zhang, Wang, Li, Guan,
  Chen, and Lin}{Xu et~al\mbox{.}}{2020}]%
        {Xu20}
\bibfield{author}{\bibinfo{person}{Pengfei Xu}, \bibinfo{person}{Xiaofan
  Zhang}, \bibinfo{person}{Cong Hao}, \bibinfo{person}{Yang Zhao},
  \bibinfo{person}{Yongan Zhang}, \bibinfo{person}{Yue Wang},
  \bibinfo{person}{Chaojian Li}, \bibinfo{person}{Zetong Guan},
  \bibinfo{person}{Deming Chen}, {and} \bibinfo{person}{Yingyan Lin}.}
  \bibinfo{year}{2020}\natexlab{}.
\newblock \showarticletitle{{AutoDNNchip: An Automated DNN Chip Predictor and
  Builder for Both FPGAs and ASICs}}.
\newblock \bibinfo{journal}{\emph{ACM/SIGDA International Symposium on
  Field-Programmable Gate Arrays}} (\bibinfo{year}{2020}),
  \bibinfo{pages}{40–50}.
\newblock


\bibitem[\protect\citeauthoryear{Zeng, Chen, Zhang, and Prasanna}{Zeng
  et~al\mbox{.}}{2018}]%
        {Zeng2018}
\bibfield{author}{\bibinfo{person}{Hanqing Zeng}, \bibinfo{person}{Ren Chen},
  \bibinfo{person}{Chi Zhang}, {and} \bibinfo{person}{Viktor Prasanna}.}
  \bibinfo{year}{2018}\natexlab{}.
\newblock \showarticletitle{{A Framework for Generating High Throughput CNN
  Implementations on FPGAs}}.
\newblock \bibinfo{journal}{\emph{ACM/SIGDA International Symposium on
  Field-Programmable Gate Arrays}} (\bibinfo{year}{2018}),
  \bibinfo{pages}{117–126}.
\newblock


\bibitem[\protect\citeauthoryear{Zhang, Wang, Zhu, Lin, Xiong, Hwu, and
  Chen}{Zhang et~al\mbox{.}}{2018}]%
        {Zhang18}
\bibfield{author}{\bibinfo{person}{Xiaofan Zhang}, \bibinfo{person}{Junsong
  Wang}, \bibinfo{person}{Chao Zhu}, \bibinfo{person}{Yonghua Lin},
  \bibinfo{person}{Jinjun Xiong}, \bibinfo{person}{Wen-mei Hwu}, {and}
  \bibinfo{person}{Deming Chen}.} \bibinfo{year}{2018}\natexlab{}.
\newblock \showarticletitle{{DNNBuilder: an Automated Tool for Building
  High-Performance DNN Hardware Accelerators for FPGAs}}.
\newblock \bibinfo{journal}{\emph{IEEE/ACM International Conference on
  Computer-Aided Design}} (\bibinfo{year}{2018}), \bibinfo{pages}{1--8}.
\newblock


\bibitem[\protect\citeauthoryear{Zhang, Ye, Wang, Lin, Xiong, Hwu, and
  Chen}{Zhang et~al\mbox{.}}{2020}]%
        {Zhang20}
\bibfield{author}{\bibinfo{person}{Xiaofan Zhang}, \bibinfo{person}{Hanchen
  Ye}, \bibinfo{person}{Junsong Wang}, \bibinfo{person}{Yonghua Lin},
  \bibinfo{person}{Jinjun Xiong}, \bibinfo{person}{Wen-mei Hwu}, {and}
  \bibinfo{person}{Deming Chen}.} \bibinfo{year}{2020}\natexlab{}.
\newblock \showarticletitle{{DNNExplorer: A Framework for Modeling and
  Exploring a Novel Paradigm of FPGA-Based DNN Accelerator}}.
\newblock \bibinfo{journal}{\emph{IEEE/ACM International Conference on
  Computer-Aided Design}} (\bibinfo{year}{2020}), \bibinfo{pages}{1--9}.
\newblock


\bibitem[\protect\citeauthoryear{Zhao, Liu, Du, Guo, Hu, Zhuang, Zhang, Song,
  Li, Zhang, Li, Xu, and Chen}{Zhao et~al\mbox{.}}{2021}]%
        {Zhao21}
\bibfield{author}{\bibinfo{person}{Yongwei Zhao}, \bibinfo{person}{Chang Liu},
  \bibinfo{person}{Zidong Du}, \bibinfo{person}{Qi Guo}, \bibinfo{person}{Xing
  Hu}, \bibinfo{person}{Yimin Zhuang}, \bibinfo{person}{Zhenxing Zhang},
  \bibinfo{person}{Xinkai Song}, \bibinfo{person}{Wei Li},
  \bibinfo{person}{Xishan Zhang}, \bibinfo{person}{Ling Li},
  \bibinfo{person}{Zhiwei Xu}, {and} \bibinfo{person}{Tianshi Chen}.}
  \bibinfo{year}{2021}\natexlab{}.
\newblock \showarticletitle{{Cambricon-Q: A Hybrid Architecture for Efficient
  Training}}.
\newblock \bibinfo{journal}{\emph{International Symposium on Computer
  Architecture}} (\bibinfo{year}{2021}), \bibinfo{pages}{706--719}.
\newblock


\end{thebibliography}

\end{document}